\documentclass[10pt,journal]{IEEEtran}
\usepackage{amsmath,amsfonts}
\usepackage{algorithmic}
\usepackage{algorithm}
\usepackage{array}
\usepackage[caption=false,font=normalsize,labelfont=sf,textfont=sf]{subfig}
\usepackage{textcomp}
\usepackage{stfloats}
\usepackage{url}
\usepackage{verbatim}
\usepackage{graphicx}
\usepackage{cite}

\usepackage{multirow}
\usepackage{makecell}
\usepackage{soul}
\usepackage{color}
\usepackage[table,dvipsnames]{xcolor} 

\definecolor{mygray}{gray}{.7}
\colorlet{myred}{Red!40}
\colorlet{myblue}{Blue!30}

\colorlet{colorFst}{Green!25}       
\colorlet{colorSnd}{SpringGreen!45} 
\colorlet{colorTrd}{Yellow!30}      
\newcommand{\fs}{\cellcolor{colorFst}}   
\newcommand{\nd}{\cellcolor{colorSnd}}      
\newcommand{\rd}{\cellcolor{colorTrd}}      

\begin{document}

\title{RSB-Pose: Robust Short-Baseline Binocular 3D Human Pose Estimation with Occlusion Handling}

\author{Xiaoyue Wan, Zhuo Chen, Yiming Bao, Xu Zhao
\thanks{Xiaoyue Wan, Zhuo Chen, and Xu Zhao are with the School of Electronic Information and Electrical Engineering, Shanghai Jiao Tong University, Shanghai, 200240, China (e-mail: sherrywaan@sjtu.edu.cn; chzh9311@sjtu.edu.cn; zhaoxu@sjtu.edu.cn).}
\thanks{Yiming Bao is with the School of Biomedical Engineering, Shanghai Jiao Tong University, Shanghai, 200240, China (e-mail: yiming.bao@sjtu.edu.cn).}
}

\markboth{Journal of \LaTeX\ Class Files,~Vol.~14, No.~8, August~2021}%
{Shell \MakeLowercase{\textit{et al.}}: A Sample Article Using IEEEtran.cls for IEEE Journals}

\IEEEpubid{0000--0000/00\$00.00~\copyright~2021 IEEE}

\maketitle

\begin{abstract} 
In the domain of 3D Human Pose Estimation, which finds widespread daily applications, the requirement for convenient acquisition equipment continues to grow. To satisfy this demand, we focus on a short-baseline binocular setup that offers both portability and a geometric measurement capability that significantly reduces depth ambiguity. However, as the binocular baseline shortens, two serious challenges emerge: first, the robustness of 3D reconstruction against 2D errors deteriorates; second, occlusion reoccurs frequently due to the limited visual differences between two views. To address the first challenge, we propose the Stereo Co-Keypoints Estimation module to improve the view consistency of 2D keypoints and enhance the 3D robustness. In this module, the disparity is utilized to represent the correspondence of binocular 2D points, and the Stereo Volume Feature (SVF) is introduced to contain binocular features across different disparities. Through the regression of SVF, two-view 2D keypoints are simultaneously estimated in a collaborative way which restricts their view consistency. Furthermore, to deal with occlusions, a Pre-trained Pose Transformer module is introduced. Through this module, 3D poses are refined by perceiving pose coherence, a representation of joint correlations. This perception is injected by the Pose Transformer network and learned through a pre-training task that recovers iterative masked joints. Comprehensive experiments on H36M and MHAD datasets validate the effectiveness of our approach in the short-baseline binocular 3D Human Pose Estimation and occlusion handling.
\end{abstract}

\begin{IEEEkeywords}
3D Human Pose Estimation, Short-Baseline Binocular, Occlusion Handling, Stereo Co-Keypoints, Pose Coherence.
\end{IEEEkeywords}

\section{Introduction}
\IEEEPARstart{3}{D} Human Pose Estimation (HPE) for single-frame image is a widely studied task in computer vision with diverse applications \cite{WANG2021103225}. There are two common input configurations: monocular and multiview, each with constraints in practical implementation. Monocular technologies \cite{pavlakos2017coarse, martinez2017simple, wu2021limb} are hampered by their inherent depth ambiguity. Meanwhile, multiview methods \cite{iskakov2019learnable, he_epipolar_2020,zhang_adafuse_2021} are limited by the demands of the laboratory environment, making them less suitable for in-the-wild expansion. Binocular setup, particularly short-baseline setting, has both the benefits of multiview geometric measurement with the portability of monocular systems. However, given the advantages, the short-baseline binocular 3D HPE has not received the deserved attention in recent research.

\begin{figure}[!t]
\centering
\includegraphics[width=0.48\textwidth]{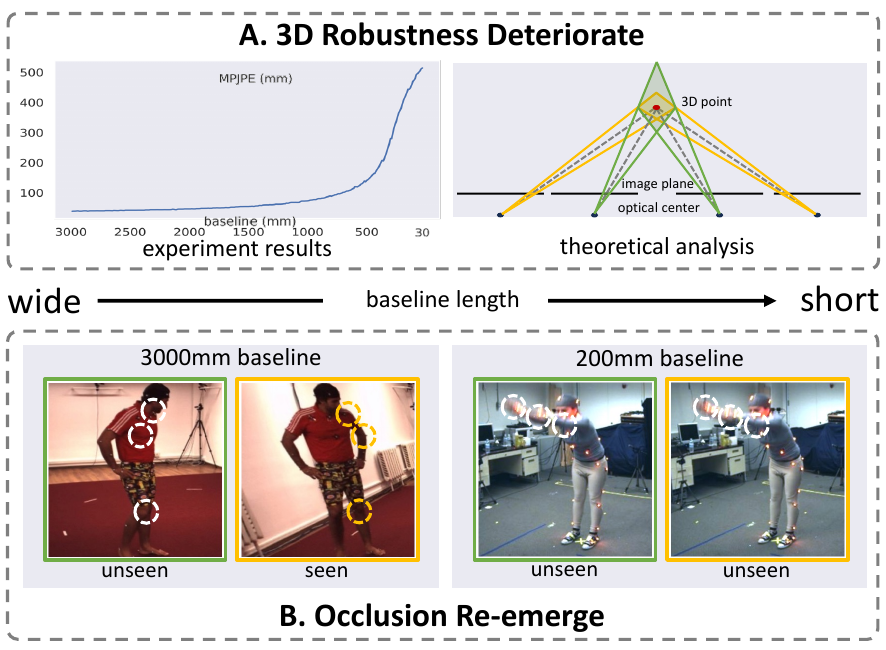}
\caption{Two main challenges of short-baseline binocular 3D human pose estimation: \textbf{A.} 3D reconstruction robustness against 2D keypoint errors deteriorates; \textbf{B.} occlusion re-emerges in both views. In A, the \sethlcolor{colorTrd}\hl{yellow} and \sethlcolor{colorFst}\hl{green} intersection zones show the horizontal tangent plane of uncertainty region under different baselines respectively. In B, the \sethlcolor{colorFst}\hl{green} boxes indicate the left image, while \sethlcolor{colorTrd}\hl{yellow} boxes represent the right one. The \textcolor{White}{\sethlcolor{mygray}\hl{white}} circles indicate the occluded point, and the \sethlcolor{colorTrd}\hl{yellow} ones are visible.}
\label{fig_problem}
\end{figure}

Similar to general multiview methods, the fundamental framework of binocular 3D HPE can also be established on the epipolar geometry\cite{hartley_zisserman_2004}. It comprises two main components: a 2D detector to predict each-view 2D keypoints, and the Triangulation method \cite{hartley_zisserman_2004} to reconstruct 3D keypoints. Nonetheless, when applying this framework to short-baseline binocular scenarios, two significant challenges arise: \textbf{the robustness of 3D reconstruction against 2D keypoint errors deteriorates with shorter baseline,} and \textbf{compared to the wider baseline scenarios in multiview, occlusions re-emerge as a problem due to the limited perspective differences in short-baseline binocular settings.} \IEEEpubidadjcol First, to explore the 3D robustness across various baseline lengths, we analyze from experimental to theoretical. Two-view 2D keypoint clusters (follow a Gaussian distribution $\mathcal{N}(0, 10)$ around groundtruth) are projected back into 3D space. The 3D error distribution, depicted in the left of Fig. \ref{fig_problem}A, emphasizes that 3D accuracy extremely decreases as the baseline decreases from $3000 mm$ to $30 mm$ under the same 2D error. From theoretical analysis, we conclude that the 3D error increases as the baseline decreases, as shown on the right side of Fig. \ref{fig_problem}A (green area vs. yellow area). Second, by visualizing binocular images under different baseline conditions, we discover that two perspectives tend to provide more visual differences in the case of a wider baseline. Consequently, occlusion occurs more frequently in two views in the short-baseline binocular scenario. For instance, as shown in Fig. \ref{fig_problem}B, the occluded left elbow in the left view becomes visible in the right view under a $3000 mm$ baseline, while the right arm is occluded in both views under a $200 mm$ baseline. 


To enhance 3D robustness against 2D errors, studies \cite{qiu_cross_2019, he_epipolar_2020, remelli2020lightweight, zhang_adafuse_2021, ma_ppt_2022} pursue the multiview consistency of 2D results, ensuring they are the projection from one target in 3D space. These approaches develop more rapidly compared to improving the 2D detector view independently. For example, \cite{qiu_cross_2019, he_epipolar_2020, zhang_adafuse_2021} enhance features or heatmaps in one view with information from other views along epipolar lines to improve view consistency. Meanwhile, \cite{remelli2020lightweight, ma_ppt_2022} employs a uniform representation to merge multiview features. However, in these methods, the information from other views serves as an auxiliary and is primarily used for feature enhancement rather than regression, resulting in limited improvement in view consistency. \textbf{The underlying reason is that a point in one view can only identify its corresponding line in another view, rather than an exact point. This limitation prevents the precise correspondence of features across multiple views, reducing their reliability and causing them to be considered auxiliary.} Hence, the main challenge is to find the exact correspondence, which \textbf{disparity} precisely reflects. In addition, a structure for keeping this corresponding relationship is also needed.

Upon the analysis, we propose the Stereo Volume Feature (SVF), a 4D structural feature that concatenates left features with their corresponding right features across various disparities. \textbf{The SVF is designed to enable binocular features with exact correspondences to jointly determine the most likely object to be the two-view 2D keypoints, rather than merely serving as an auxiliary feature enhancement.} After regression, a co-heatmap is generated. This is a 3D probability heatmap whose value represents the probability of each grid in SVF to be the target, while the localization indicates the left-view 2D position and its disparity to the right-view point, named co-keypoint. Through this collaborative regression, the view consistency is effectively restricted to binocular 2D keypoints, while the extended disparity dimension also allows for increased consideration of the more ambiguous depth axis. Furthermore, the disparity formulation forces binocular keypoints to share the same Y-localization, effectively leveraging the epipolar constraint.
Additionally, an Attention Mask (AM) is introduced to filter out perturbed features in each view, thereby facilitating the convergence of the SVF regression. Combining AM and SVF, our novel 2D keypoint estimation module, Stereo Co-Keypoints Estimation (SCE), is proposed.

The occlusion problem originates from the fact that the additional visual complementary provided by the other view is quite limited due to the short baseline. \textbf{Intuitively, injecting pose coherence, i.e., modeling semantic information within a 3D pose like joint correlations, can guide the occluded joints from other visible joints.} Recent works have harnessed Transformer-based methods \cite{zhang_mixste_2022, tang_3d_nodate,zhao_poseformerv2_2023, xue2022boosting, wu2021limb} to consider spatial dependencies among joints in multi-frame tasks. These studies demonstrate the capability of the Transformer to capture joint correlations. However, most approaches primarily focus on enhancing the temporal smoothness of these correlations in 2D feature extraction. Few works employ the Transformer to directly capture the semantic information within the 3D pose, and this is the objective of Pose Transformer (PT) in our research. \textbf{To actuate PT for extracting pose coherence more effectively, we design a self-supervised pre-training task involving recovering masked joints, which is inspired by Bert \cite{devlin2018bert}.} Following this, the pre-trained PT (PPT) is integrated into the entire framework to refine the initial 3D poses reconstructed via Triangulation and make them perceive pose coherence. Furthermore, to bridge the input distribution gap between pre-training groundtruth and inference estimation, we introduce an iterative masking strategy during pre-training, allowing simultaneous data augmentation. 

Our method, named RSB-Pose, is constructed by integrating the basic framework with SCE and PPT modules.
It is trained end-to-end and rigorously evaluated on two datasets: H36M, which represents wide-baseline binocular scenarios, and MHAD, which is short-baseline binocular settings. The results show that RSB-Pose is competitive with state-of-the-art methods, particularly excelling in short-baseline binocular scenarios. Additionally, experiments on the MHAD\_occ dataset highlight the occlusion-handling capability of PPT and validate the effectiveness of RSB-Pose. The contributions of this work can be summarized as follows:

\begin{itemize}
    \item{We propose a novel SCE module for estimating binocular 2D keypoints, which improves view consistency using SVF. Leveraging disparity, SVF offers a more efficient and flexible representation of binocular point correspondences. SCE module improves the robustness of 3D reconstruction, particularly in short-baseline scenarios.}
    \item{We introduce the PPT to enhance 3D pose coherence and address frequent occlusion scenarios in short-baseline effectively. The pre-training strategy enables PT to capture semantic information within the 3D pose.}
    \item{Our RSB-Pose method significantly enhances state-of-the-art performance on both H36M and MHAD datasets. A comprehensive set of experiments are conducted to demonstrate the effectiveness of our approach.}
\end{itemize}

\begin{figure*}[!t]
\centering
\includegraphics[width=\textwidth]{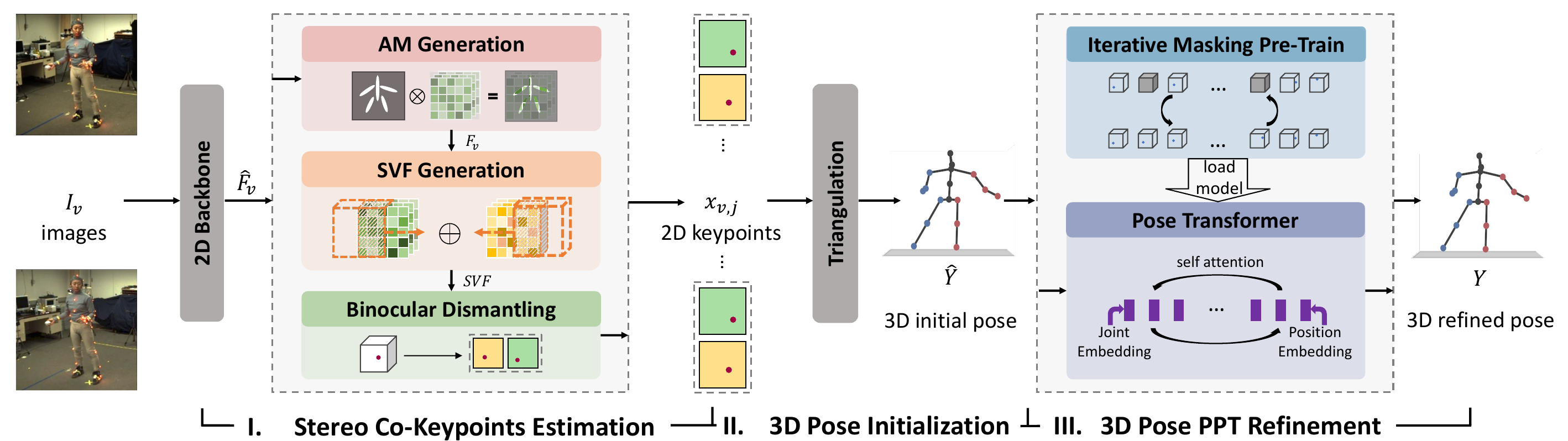}
\caption{The framework of RSB-Pose. The binocular images are firstly encoded by a 2D backbone and then processed through three main steps: \textbf{\uppercase\expandafter{\romannumeral1}.} Stereo Co-Keypoints Generation: Two-view features are concatenated in the Stereo Volume Feature (SVF), facilitating the simultaneous regression of 2D binocular keypoints and ensuring their view consistency; \textbf{\uppercase\expandafter{\romannumeral2}.} 3D Pose Initialization: Triangulation is utilized to reconstruct the initial 3D pose; \textbf{\uppercase\expandafter{\romannumeral3}.} 3D Pose PPT Refinement: Pose coherence is perceived by the Pose Transformer through pre-training and then injected into the refined 3D pose.}
\label{fig_framework}
\end{figure*}

\section{RELATED WORK}

\subsection{Monocular 3D Human Pose Estimation}
\noindent Monocular 3D HPE focuses on predicting the human pose in three-dimensional space using a single-view image as input. Previous works can be broadly categorized into two main approaches: one-stage and two-stage. One-stage methods \cite{pavlakos2017coarse, tome2017lifting, zhou2017towards, tekin2016structured, tekin2016direct, tu2023consistent} rely on extensive image-pose pair datasets and carefully designed network architectures to improve performance. On the contrary, two-stage methods \cite{martinez2017simple, kocabas2019self, chen2019weakly,Kocabas_2020_CVPR, zeng2021learning, zhang2022uncertainty, zhou2023dual} employ off-the-shelf 2D detectors \cite{newell2016stacked, sun2019deep, kan2023self} to initially estimate the 2D pose from the image. Subsequently, various network structures such as fully connected networks, graph convolution networks, or Transformer networks are utilized to lift the 2D pose to the corresponding 3D pose. Despite the incorporation of geometric constraints \cite{rhodin2018unsupervised}, and human models \cite{li2021hybrik}, monocular methods still suffer from the inherent challenges of depth ambiguity.

\subsection{Binocular and Multiview 3D Human Pose Estimation}
\noindent Currently, there are few methods designed specifically for binocular 3D HPE. Binocular settings are usually found in the evaluation of view number within Multiview studies. Therefore, we merge binocular and multiview methods in this section. These approaches leverage view geometric constraints \cite{hartley_zisserman_2004} to address the depth ambiguity encountered in monocular, shifting the task from regression to a measurement-based manner. The fundamental framework consists of two steps: predict 2D features, heatmaps, or keypoints, and reconstruct 3D from 2D cues. Based on the strategy of 2D-3D, multiview methods can be categorized into two streams: model-based and model-free. In model-based methods \cite{burenius20133d,pavlakos2017harvesting,qiu_cross_2019}, a 3D model serves as the optimization objective. These methods optimize the 3D pose to ensure its projection align with the observed 2D cues. Model-free methods \cite{iskakov2019learnable,he_epipolar_2020,zhang_adafuse_2021,remelli2020lightweight,zhuo2022structural,WAN2023103830} rely on epipolar constraints and employ Triangulation \cite{hartley_zisserman_2004} to solve the 3D keypoint from multiview 2D keypoints by optimizing reprojection error. These methods are increasingly popular due to the mature 2D detectors and the elegant Triangulation. Therefore, we choose the model-free method as our binocular framework.

However, considering the short-baseline setting, the accuracy of multiview 2D keypoints becomes critical. Several works \cite{qiu_cross_2019, he_epipolar_2020, zhang_adafuse_2021, ma_ppt_2022, remelli2020lightweight} have explored multiview 2D detector mechanism. The primary motivation is to augment the features in one view by fusing the features from other views along its epipolar line, thus enabling 2D keypoint gain to 3D perception. However, even after such feature fusion, the regression of keypoints remains independent, which limits the effectiveness of restricting view constraints. The greatest challenge lies in the inability to guarantee the correspondence of pixels between the two views.
To address this challenge, our SCE module constructs an SVF by aggregating binocular features across various disparities. Then, it regresses co-keypoints which is a kind of 3D point that contains the locations of left-view keypoints and their corresponding keypoints in the right view. Through SCE, the geometric correspondence of the binocular keypoints can be ensured.

\subsection{Occlusion Handling}
\noindent Due to the lack of visual information during occlusion, 2D keypoints are frequently unreliable.
Prior research has explored various methods to restrict the 3D pose space to handle occlusions. These methods include using Autoencoder \cite{tekin2016structured} to map joints into latent representations, employing Generative Adversarial Networks \cite{wandt2019repnet} to model pose distributions, and applying Graph Convolutional Networks \cite{10179252, 9535240, 9009459} to capture joint correlations. More recently, Transformer-based approaches \cite{zhang_mixste_2022, tang_3d_nodate,zhao_poseformerv2_2023, xue2022boosting, wu2021limb} have been used to establish spatial-temporal dependencies among joints in multi-frame tasks. In this work, we leverage the Transformer to specifically model spatial correlations within poses because of its flexible capability to capture global correlations between nodes.

\subsection{Pre-Training of Transformer}
\noindent The remarkable success of Transformer \cite{vaswani2017attention} in NLP and CV can be largely attributed to the use of pre-training techniques \cite{yosinski2014transferable, devlin2018bert, yan2022crossloc, chang2022maskgit}. In NLP, the introduction of self-supervised tasks, such as recovering masked words, as seen in BERT \cite{devlin2018bert}, enables the model capable of capturing contextual semantic information. Similarly, in the field of CV \cite{chang2022maskgit}, a similar pre-training strategy is adopted. Pre-training is a powerful approach because self-supervised tasks efficiently expand the dataset, which is crucial for the substantial amount of training data that Transformer requires. In HPE, P-STMO \cite{shan2022p} introduced a task focused on recovering masked 2D poses to augment the training data. In this work, we propose a similar self-supervised task, but with differences in input and objective. Specifically, our task involves recovering masked 3D poses, with the goal of facilitating the exploration of spatial correlations among joints within a pose. 

\begin{figure*}[!t]
\centering
\includegraphics[width=\textwidth]{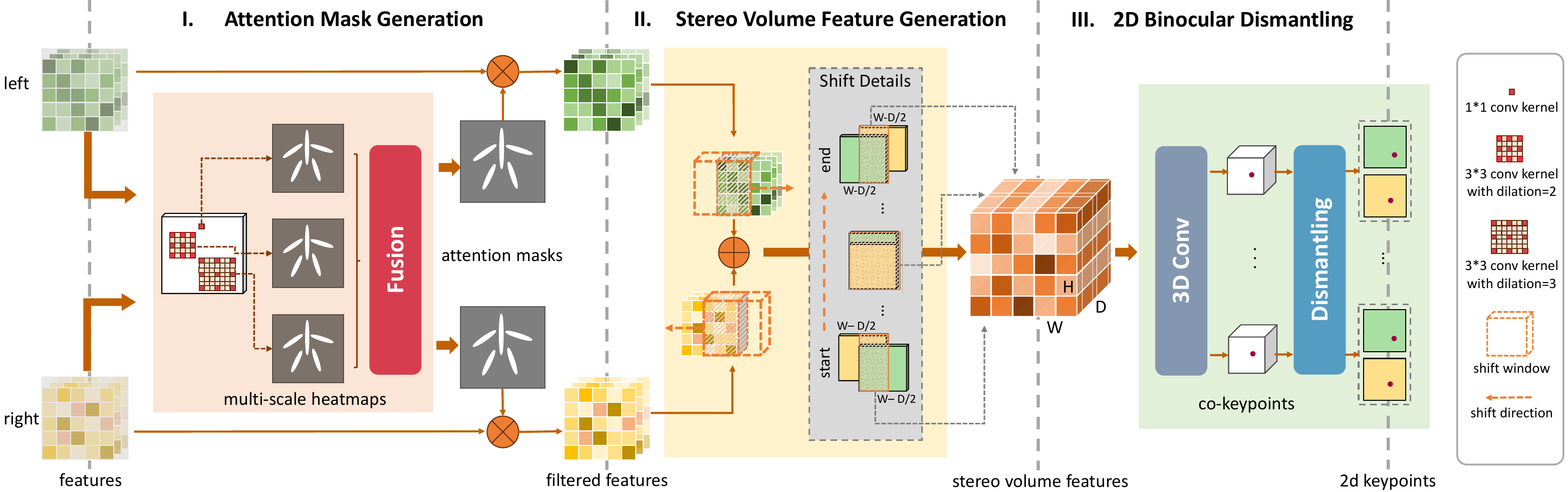}
\caption{The framework of Stereo Co-Keypoints Estimation module: \textbf{\uppercase\expandafter{\romannumeral1}.} Attention Mask Generation, to focus initial features on the huaman body of interest; \textbf{\uppercase\expandafter{\romannumeral2}.} Stereo Volume Feature Generation, to consider both binocular views simultaneously and form as a 4D feature volume; \textbf{\uppercase\expandafter{\romannumeral3}.} 2D Binocular Dismantling, to solve binocular 2D keypoints from co-keypoints regressed from SVF.}
\label{fig_stereo_co_heatmap}
\end{figure*}

\section{METHODOLOGY}

\subsection{Framework}
\noindent Our RSB-Pose framework is illustrated in Fig. \ref{fig_framework}. It comprises three primary steps: SCE, 3D Pose Initialization, and 3D Pose PPT Refinement. The model takes single-frame binocular images $I_v$ as input, which have been rectified using the Stereo Rectification method \cite{loop_computing_1999} and cropped according to the groundtruth bounding box. Here, $v\in \{0,1\}$ signifies the two views, with the index $0$ referring to the left view. By convention, an off-the-shelf 2D backbone is utilized to extract initial features $\hat{\mathbf{F}}_v\in \mathbb{R}^{(C,H,W)}$ of each view. Then the SCE module is utilized to estimate the co-keypoints. 

In SCE, there are three main modules: AM Generation, SVF Generation, and 2D Binocular Dismantling. Features $\hat{\mathbf{F}}_v$ are first down-sampled to 16 dimensions by a $1 \times 1$ convolution layer. Meanwhile, the AM module generates an attention mask $\mathbf{M}_v\in \mathbb{R}^{(H,W)}$ to emphasize anatomical parts within $\hat{\mathbf{F}}_v$. Then, by concatenating the filtered binocular features $\mathbf{F}_v$, $\mathbf{SVF}\in\mathbb{R}^{(32,D,H,W)}$ is formulated. $D$ denotes the dimension of the disparity, where each $\mathbf{SVF}(d,h,w)$ represents the co-features of the left 2D point $(h,w)$ and its corresponding right point under disparity $d$. A 3D convolution network is used to regress co-keypoints, which are then dismantled into binocular 2D keypoints $\mathbf{x}_{v,j}\in\mathbb{R}^{2}$, $j\in[0, J)$ is the index of keypoints. The skeleton here has $J=17$ keypoints. 

In the 3D Pose Initialization stage, 3D keypoints $\mathbf{y}_{v,j}\in\mathbb{R}^{3}$ are reconstructed individually using Triangulation \cite{hartley_zisserman_2004} and then concatenated into a 3D pose $\hat{\mathbf{Y}}\in\mathbb{R}^{(J,3)}$. 

Finally, to refine the 3D pose, 3D Pose PPT Refinement stage is added to inject overall pose coherence. In which, the PPT takes the initial 3D pose as input and produces the refined 3D pose $\mathbf{Y}\in\mathbb{R}^{(J,3)}$. The pose coherence is learned by a self-supervision pre-training task. In the next sections, SCE and PPT modules will be described in detail.

\subsection{Stereo Co-Keypoints Estimation}
\noindent Several works \cite{zhang_adafuse_2021, ma_ppt_2022, qiu_cross_2019, he_epipolar_2020} have leveraged multiview constraints to improve 2D keypoints view consistency. However, a prevalent approach enhances the single-view features of a pixel by utilizing features from another view along its epipolar line and regresses the pixel probability to identify the keypoint. Essentially, during the regression step, the 2D keypoint remains independent of another view, with limited consideration given to auxiliary-view features, the geometric correspondence of which even cannot be guaranteed. The view consistency constraints are finally not fully leveraged. We agree that the likelihood of a pixel in one view being the keypoint should be jointly determined by its corresponding pixel in another view. But the central challenge lies in identifying this corresponding pixel. Here, we draw inspiration from Stereo Matching \cite{xu_aanet_2020, xu_attention_2022, kendall_end--end_2017} and utilize disparity to describe the corresponding relationship, which is then expressed in the SVF. After regressing the co-heatmaps which is a 3D probabilistic map that represents each grid in SVF as a potential keypoint, the co-keypoints for both the left and right views can be generated simultaneously. 


\subsubsection{Stereo Volume Feature}
\noindent In Stereo Matching, a cost volume with dimensions $D\times H\times W$ is generated to depict the matching degree between two 2D binocular points at each disparity level. The grid with the highest probability signifies that the two binocular counterparts match and are projected from one 3D point. Taking inspiration from this, we generate SVF through the concatenation of features from the left view with their corresponding features at varying disparity levels in the right view. Each grid feature serves as a co-feature of two-view 2D points and a representation for a 3D point situated on the ray that originates from the left 2D point simultaneously.

In Stereo Matching, the cost volume is typically generated by concatenating binocular features cropped by a shift window. The window shifts along a fixed direction for each view. Specifically, the shift window in the left view starts from cropping the whole image and shifts along the positive direction of W axis, while the right view window shifts along the opposite direction but from the same starting point. The shifting method is appropriate because, for one 3D point, the horizontal position of its right-view point is consistently smaller than that of its left-view point. But this differs from our SVF, where the bounding box cropping does not maintain this relationship. Therefore, we adapt the feature generation formulation to accommodate the bidirectional scenario. In particular, as depicted in Fig. \ref{fig_stereo_co_heatmap}\uppercase\expandafter{\romannumeral2}, rather than altering the direction of shifting, we transform the starting and ending points of the shift window in each view. In the left view, the starting point of the right border line is transformed from $w=W$ to $w=W-\frac{D}{2}$ and the endpoint of the left border line is modified to $w=W-\frac{D}{2}$. In the right view, the left border line starts at $w=W-\frac{D}{2}$ and the right border line ends at $w=W-\frac{D}{2}$. The correspondence of index to reality in the disparity dimension of the formulation is also adjusted. Concurrently, with the binocular window features concatenated, the SVF is then modified as follows:

\begin{equation}
    \begin{aligned}
        \label{eq_svf}
        \mathbf{SVF}(d,h,w) = Concat\{\mathbf{F}_0(h,w), \mathbf{F}_1(h,w-\hat{d})\},
        \\
        \{\hat{d} = d-D/2 \, | \, -D/2 \leq \hat{d} \leq D/2 \},
        \\
        \{d\in \mathbb{R} \, | \, 0 \leq d \leq D \}, 
    \end{aligned}
\end{equation}

\noindent where $D+1$ represents the disparity range (for convenience, D is used elsewhere), $d$, denotes the grid index along the axis of disparity, and $\hat{d}$ signifies the actual disparity. When $\hat{d} < 0$, the corresponding pixel in the right view is relatively located to the right of the left view pixel. If the width of cropped features is less than $W$, the remainder is padded with zero.

\subsubsection{Attention Mask Generation}
\noindent The SVF works for 3D grid probability regression and co-keypoints generation. However, the target grid is often quite sparse throughout the entire volume, which hinders the regression task. To address this issue, the attention mask $\mathbf{M}_v$ is designed to emphasize binocular features within the regions of anatomy key-parts, thereby serving as an initial filter to eliminate interference from the background or irrelevant body parts:

\begin{equation}
    \label{eq_amg}
    \mathbf{F}_v = \mathbf{M}_v \odot \hat{\mathbf{F}}_v,
\end{equation}
where $\odot$ is the element-wise product.

As shown in Fig. \ref{fig_stereo_co_heatmap}\uppercase\expandafter{\romannumeral1}, the AM generation network comprises three multi-scale heatmap regression modules and one fusion module. To extract heatmaps that account for multi-scale receptive fields, we employ a $1 \times 1$ convolution layer to derive pixel-dimensional heatmaps and utilize two $3 \times 3$ convolution layers with different dilation rates to capture likelihoods associated with larger receptive regions to distinct background and human. In detail, the dilation rates are set as $2$ and $3$ to separately capture thin and wide parts of the body structure. At the end, a $1 \times 1$ convolution layer is used to fuse these heatmaps and generate the final mask.

\begin{figure}[!t]
\centering
\includegraphics[width=0.98\linewidth]{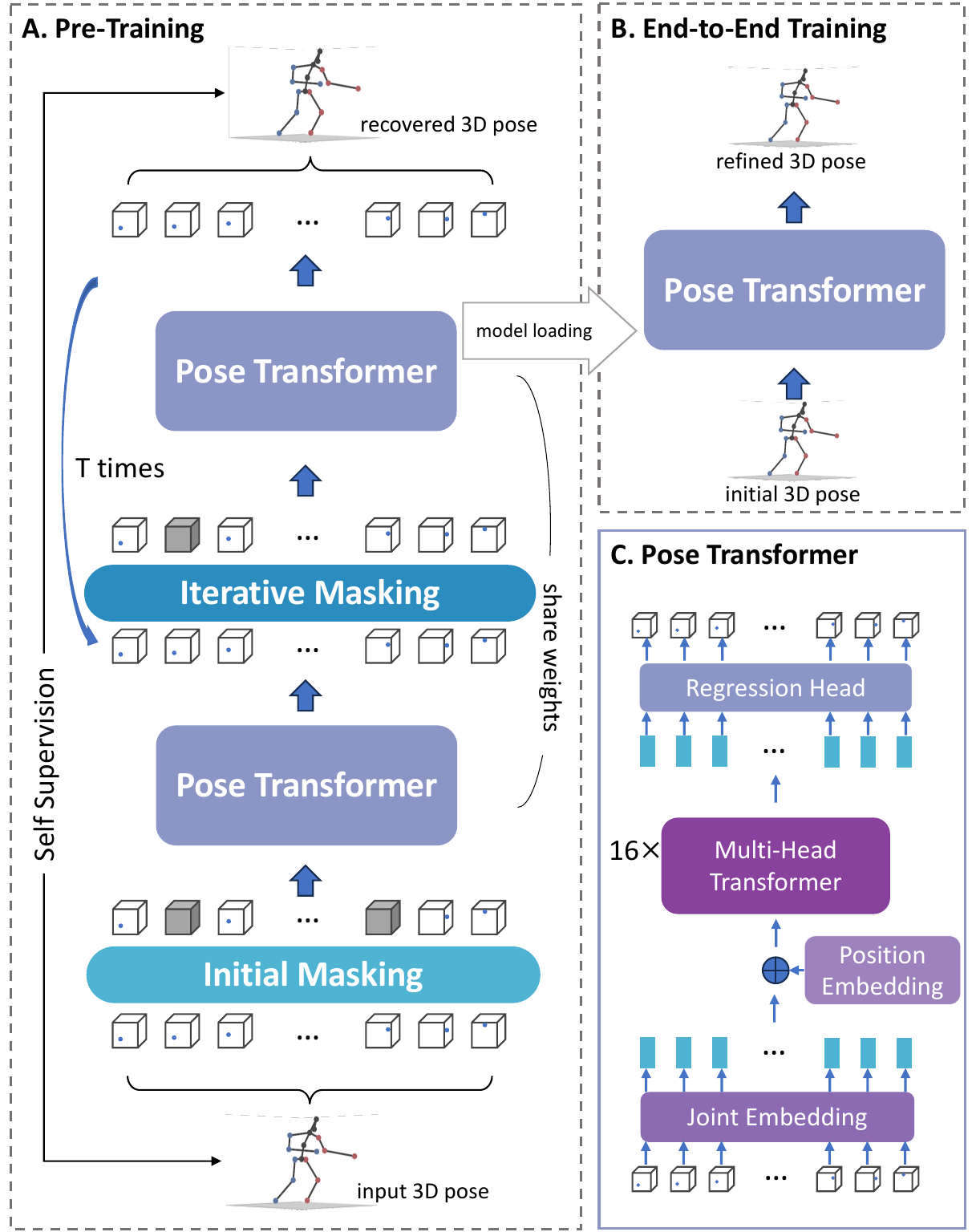}
\caption{Illustration of Pre-trained Pose Transformer: \textbf{A.} Pre-Training Strategy, \textbf{B.} End-to-End Training within the framework, \textbf{C.} Pose Transformer Structure. Firstly, the Pose Transformer undergoes a pre-training stage with a self-supervised task, involving iterative recovery of masked poses. Subsequently, during the whole framework end-to-end training, the Pre-trained Pose Transformer is reloaded and receives initial predicted 3D poses as input.}
\label{fig_pre-trained_pose_transformer}
\end{figure}

\subsubsection{2D Binocular Dismantling}
\noindent As depicted in Fig. \ref{fig_stereo_co_heatmap}\uppercase\expandafter{\romannumeral3}, the co-heatmap $\mathbf{CH}_j \in \mathbb{R}^{(D,H,W)}$ is regressed after a 3D convolution network. Applying soft-argmax \cite{sun2018integral}, the mass localization can be calculated which represents the most confident co-keypoints to be the 3D human keypoints. And according to Eq. \eqref{eq_svf}, the 2D binocular keypoints can be dismantled:

\begin{equation}
    \begin{aligned}
        \label{eq_2dkpt_dismantle}
        (d,h,w)_j &= Soft\text{-}argmax(\mathbf{CH}_j),
        \\ \mathbf{x}_{0,j} &= [h,w]^T,
        \\ \mathbf{x}_{1,j} &= [h, w-d+D/2]^T.
    \end{aligned}
\end{equation}

The SCE module can be combined with Triangulation to reconstruct 3D poses and trained independently from the entire framework. The training loss accounts for the prediction error of 3D keypoints and the error in the SVF mass location.

\begin{equation}
    \begin{aligned}
        \label{eq_preloss}
        L_{SCE} &= L_{3D} + \beta L_{SVF},
        \\L_{3D} &= \frac{1}{J}\sum_{j}||\mathbf{y}_{j} - \mathbf{y^g}_{j}||_1,
        \\L_{SVF} &= \frac{1}{J}\sum_{j}-\log (\mathbf{CH}_j(\sim{\mathbf{y^g}}_j)).
    \end{aligned}
\end{equation}
Here, $\mathbf{y}_j$ represents the estimated 3D keypoints, $\mathbf{y^g}_j \in \mathbb{R}^3$ corresponds to the groundtruth, $\sim{\mathbf{y^g}}_j$ is the nearest grid localization of the co-keypoint groundtruth in disparity space, and $\beta = 0.01$ by empirical results. The loss function $L_{3D}$ is computed as the mean per joint L1 distance between the predicted 3D keypoints and the groundtruth. $L_{SVF}$, on the other hand, draws inspiration from \cite{iskakov2019learnable} and is designed to maximize the probability in the 3D groundtruth of the co-heatmap. This ensures that the most probable grid in the co-heatmap is close to the true 3D position.

\subsection{Pre-trained Pose Transformer}
\noindent Another significant issue in short-baseline binocular settings is that the additional information provided by the other view is quite limited due to the small perspective difference between two cameras. The occluded parts in the left view are typically also occluded in the right view. Nonetheless, Triangulation cannot resolve this problem due to the separate reconstruction within 3D keypoints. Hence, we introduce PPT to refine the 3D pose considering pose coherence. Recently, some researches \cite{zhang_mixste_2022, zhao_poseformerv2_2023, tang_3d_nodate} have utilized Transformer to model spatial correlations between joints in 2D pose sequences for 3D frame pose estimation and to enforce their temporal smoothness. However, there is limited work that directly employs Transformer to model the 3D joint dependencies which represents pose coherence more intuitively. We employ PT to capture the coherence within the 3D pose and refine the initial results.

The 3D Pose Refinement process consists of two stages as illustrated in Fig. \ref{fig_pre-trained_pose_transformer}. In the first stage, PT undergoes pre-training via a self-supervised task, recovering masked keypoints within the entire 3D pose. 
Here, PT is guided to perceive the spatial correlations between joints through pre-training, which will be demonstrated in the ablation study. In the second stage, the PPT is integrated into the framework and undergoes end-to-end training.

\begin{algorithm}[!b]
\caption{Iterative Masking}
\label{alg_iterativemask}
\begin{algorithmic}
    \ENSURE $\mathbf{Y}^{g}$ input 3D pose 
    \REQUIRE $\mathbf{Y}^{r}$ recovered 3D pose
    \STATE Initial Masking: $\mathbf{Y}^{m} \gets \{\mathbf{y^{g}}_j, \, j\notin \mathbb{M}\} \cup \{\mathbf{m}, \, j\in \mathbb{M}\}$
    \STATE $iter \gets 1$
    \WHILE{$iter\, < \, T$}
    \STATE Recovery: \quad $\mathbf{Y}^{r}, \mathbf{A}^{h} \gets PT(\mathbf{Y}^{m})$
    \STATE Top-K Confident: \quad $\mathbb{K} \gets \{ topK \quad conf(\mathbf{y^r}_j), j\in \mathbb{M}\}$
    \STATE Iterative Masking: 
    \STATE \quad $\mathbf{Y}^{m} \gets \{\mathbf{y^r}_j, \, j\notin \mathbb{M}\} \cup \{\mathbf{y^r}_j, \, j\in \mathbb{K}\} \cup \{\mathbf{m}, \, j\in \mathbb{M} /\mathbb{K}\} $
    \STATE \quad $\mathbb{M} \gets \mathbb{M} /\mathbb{K}$
    \ENDWHILE
    \STATE $\mathbf{Y}^{r}, \mathbf{A}^{h} \gets PT(\mathbf{Y}^{m})$
\end{algorithmic}
\end{algorithm}

\subsubsection{Pre-Training Strategy}
\noindent  As illustrated in Fig. \ref{fig_pre-trained_pose_transformer}A, the input to the PT is a masked 3D pose denoted as $\mathbf{Y}^{m}$. Within $\mathbf{Y}^{m}$, a portion of joints belonging to the masked set $\mathbb{M}$ is substituted with a constant padding joint $\mathbf{m}\in \mathbb{R}^3$. The objective of the self-supervision task is to reconstruct the original 3D pose $\mathbf{Y}^{g}$. During pre-training, the 3D pose input is from groundtruth, but during inference, it will be the predicted one with a different error distribution. To address this issue, an iterative recovery strategy is employed. Specifically, as depicted in Algo. \ref{alg_iterativemask}, we perform recovery for $T$ times. At the end of each iteration, we retain the top-K confident recovered 3D points $\mathbb{K} \subset \mathbb{M}$, while the others are replaced with $\mathbf{m}$ once more. Therefore, the input for the recovery is adjusted to incorporate not only groundtruth but also previously recovered points. According to the ablation study in Sec. \ref{sec:effect_of_pretrain}, the mask ratio and $T$ are set to 0.4 and 2, respectively. Then, $K$ is set to 2 to ensure that the input of the $2^{nd}$ masking iteration still contains masked joints. The confidence of one recovered point $\mathbf{y^r}_j$ is calculated as the sum of the attention weights other joints query it:

\begin{equation}
    \label{eq_recoverconf}
    conf(\mathbf{y^r}_j) = \sum_{a} \sum_{h} \mathbf{A}^h(a,b=j),
\end{equation}
where $\mathbf{A}^h\in \mathbb{R}^{(J, J)}$ is the multi-head attention map in the final layer of PT, and $h$ is the index of attention heads. How much attention the joint $a$ pays to the joint $b$ is described by $\mathbf{A}^h(a,b)$. The top-K confident points are selected under the hypothesis that the most confident keypoint should have stronger connections to other keypoints compared to the less confident ones.

\subsubsection{Pose Transformer}
The PT module is designed as shown in Fig. \ref{fig_pre-trained_pose_transformer}C, which is similar to the Spacial Transformer module of \cite{zheng20213d}. Given a 3D pose, we consider the pose as $J$ separate joints. To enhance the representation of these 3D keypoints, a linear embedding layer is utilized to transform the spatial locations of the joints into high-dimensional feature vectors. Additionally, we embed the cross-joint positional relationships using learnable parameters. These joint features, denoted as $\mathbf{E}\in \mathbb{R}^{J \times dim_{e}}$, and joint positions, denoted as $\mathbf{E}_{P}\in \mathbb{R}^{J \times dim_{e}}$, are concatenated and fed into the pose encoder, where $dim_{e} = 128$ as determined by ablation study in Sec. \ref{sec:appendix_impact_of_PT_network} of Appendix. The pose encoder is constructed by stacking 16 Multi-Head Transformer Encoders \cite{vaswani2017attention}. Each encoder consists of a multi-head attention layer followed by a multi-layer perception layer. LayerNorm is employed both before and after the attention layer. Within the attention layer, we utilize 8 attention heads and apply the scaled dot-product attention mechanism to calculate the attention maps $\mathbf{A}^h$. Finally, a regression head, implemented as a linear layer, is employed to generate the final refined 3D pose.

The loss functions of self-supervised pre-training and end-to-end whole framework training are both 3D MPJPE loss:

\begin{equation}
    \label{eq_mpjpeloss}
    L_{MPJPE} =  \frac{1}{J}  \sum_{j} ||\mathbf{y}_{j} - \mathbf{y^g}_{j}||_2.
\end{equation}

\section{EXPERIMENTS}
\subsection{Datasets and Experimental Settings}
\subsubsection{Datasets}
\noindent The validations are conducted on MHAD Berkeley dataset \cite{ofli2013berkeley} and the H36M dataset \cite{ionescu2013human3}, 
representative of binocular scenarios with short baseline and wide baseline, respectively.

MHAD is a multi-modal dataset that encompasses 11 actions performed by 12 subjects. The data acquisition system includes multiview stereo vision camera arrays, Kinect cameras, wireless accelerometers, etc. To assess our performance in the short-baseline binocular setting, we opt for two couple cameras in the L1 quad camera, $1^{st}$ and $3^{th}$, $2^{nd}$ and $4^{th}$, which have an approximate baseline of 200 mm. Similar to previous work \cite{makris_robust_2019, ying_rgb-d_2021}, subjects 8 and 11 are used for testing.

H36M is the most popular dataset for 3D human pose estimation. This dataset boasts 3.6 million annotations, covering a wide range of scenarios performed by 11 different actors. For training, subjects 1, 5, 6, 7, and 8 are used, while subjects 9 and 11 are reserved for testing. The videos are captured using 4 synchronized high-resolution cameras placed around the laboratory. We select the $2^{nd}$ and $4^{th}$, $1^{st}$ and $3^{th}$ camera pairs to provide a baseline setting of approximately 3000 mm, which is applied to evaluate the generalized wide-baseline binocular performance of our method.

In addition, we filter out the occluded joints in the MHAD, resulting in a subset named MHAD\_occ, for validating the occlusion handling of PPT. With 2D groundtruth keypoints available, we identify the occluded joints based on their distance from other skeleton bones. The distance threshold varies depending on the bone type, with thinner arm bones having a smaller threshold and thicker torso bones having a larger one. Furthermore, we conduct a manual inspection to ensure the accuracy of the occlusion identification.

\subsubsection{Evaluation Metrics}
\noindent Two metrics, Joint Detection Rate (JDR) and Mean Per Joint Position Error (MPJPE), are utilized to assess the accuracy of 2D and 3D estimations, respectively. Regarding the JDR metric, a joint is successfully detected if its distance to the groundtruth is smaller than half of the head size. Since the head size is not provided, we set it to be $2.5\%$ of the bounding box width \cite{zhang_adafuse_2021}. To validate 3D pose, we develop two MPJPE metrics: MPJPE\_ab, which measures the error in predicted absolute joint locations, and MPJPE\_re, which quantifies the error in keypoints relative to the pelvis.

To validate the view consistency of binocular 2D keypoints, we employ the cosine similarity of features from two views cropped at each 2D keypoint location, termed SIM\_cos. The features are definite as the last layer of features before the regression head in the 2D backbone. The SIM\_cos ranges from $-1$ to $1$, with $1$ indicating maximum similarity. Since we assert that consistent binocular 2D keypoints are projections from the same 3D point, the cosine similarity of their features should be close to $1$, as they essentially capture the same target.

To evaluate the efficiency of methods, Multiply-Accumulate Operations (MACs) and Parameters (Params) are utilized. They are measured by thop \footnote{https://github.com/Lyken17/pytorch-OpCounter}.

\linespread{1.4}
\begin{table*}[t]
\caption{\textbf{Quantitative Comparison with SOTA methods on MHAD datasets.} The best results within the same scale are highlighted in \textbf{bolded}, and the second best is \underline{underlined}. Our baseline and RSB-Pose are emphasized by \sethlcolor{colorSnd}\hl{light green} and \sethlcolor{colorFst}\hl{green}, respectively.}
\label{tb:compare_with_sota_mhad}
\centering
\resizebox{\linewidth}{!}{
\begin{tabular}{l|l|l|c|cc|cc|cc}
\Xhline{0.12em}
\multicolumn{1}{c|}{\multirow{2}{*}{Method}} & \multicolumn{1}{c|}{\multirow{2}{*}{Venue}} & \multicolumn{1}{c|}{\multirow{2}{*}{2D Backbone}} & \multirow{2}{*}{Input Size} & \multirow{2}{*}{MACs (G)} & \multirow{2}{*}{Params (M)} & \multicolumn{2}{c|}{3D Pose}   & \multicolumn{2}{c}{2D Pose} \\
\multicolumn{1}{c|}{}                        & \multicolumn{1}{c|}{}                       & \multicolumn{1}{c|}{}                             &                             &                          &                             & MPJPE\_re (mm) $\downarrow$ & MPJPE\_ab (mm) $\downarrow$& JDR (\%) $\uparrow$     & SIM\_cos  $\uparrow$    \\
\hline
TPPT  \cite{ma_ppt_2022}                                       & ECCV'22                                    & TokenPose                                        & 256$\times$256                & 9.70                     & 7.75                        & 225.97        & 209.03        & 87.67        & 0.91         \\
Epipolar-T.    \cite{he_epipolar_2020}                              & CVPR'20                                    & ResNet-50                                        & 256$\times$256                    & 52.28                    & 34.07                       & 91.25         & 90.73         & 91.75        & 0.94         \\ 
\rowcolor{colorSnd}HRNet + T.   \cite{sun2019deep}                               & CVPR'19                                    & HRNet-W32                                        & 256$\times$256                     & 20.68                    & 28.55                       & 78.10         & 73.63         & 96.27        & 0.91         \\ 
\rowcolor{colorFst} RSB-Pose-HRNet                              &                                            & HRNet-W32                                        & 256$\times$256                     & 59.40                    & 37.70                       & \underline{31.10} \scriptsize{\textcolor{red}{47.00$\downarrow$}}        & \underline{35.17} \scriptsize{\textcolor{red}{38.46$\downarrow$}}         & \textbf{98.81} \scriptsize{\textcolor{red}{2.54$\uparrow$}}       & \textbf{0.99} \scriptsize{\textcolor{red}{0.08$\uparrow$}}        \\
\rowcolor{colorSnd}ResNet + T.   \cite{xiao2018simple}                               & ECCV'18                                    & ResNet-50                                        &  256$\times$256                      & 25.87                    & 33.99                       & 62.70         & 63.18         & 96.03        & 0.84         \\
\rowcolor{colorFst} RSB-Pose-ResNet50                          &                                            & ResNet-50                                        & 256$\times$256                & 65.29                    & 43.24                       & \textbf{26.93} \scriptsize{\textcolor{red}{35.77$\downarrow$}}        & \textbf{32.10} \scriptsize{\textcolor{red}{31.08$\downarrow$}}        & \underline{96.62} \scriptsize{\textcolor{red}{0.59$\uparrow$}}       & \underline{0.96} \scriptsize{\textcolor{red}{0.12$\uparrow$}}        \\ \hline
AdaFuse    \cite{zhang_adafuse_2021}                                 & IJCV'20                                    & ResNet-152                                       & 384$\times$384                  & 149.48                   & 69.66                       & 188.80        & 189.16        & 83.46        & 0.93         \\
Algebraic-T.   \cite{iskakov2019learnable}                             & ICCV'19                                    & ResNet-152                                       & 384$\times$384                  & 104.92                   & 79.52                       & 52.28         & 51.69         & 95.95        & 0.88         \\
Volume-T.    \cite{iskakov2019learnable}                               & ICCV'19                                    & ResNet-152                                       & 384$\times$384                   & 257.15                   & 80.59                       & 28.37         & 34.67         & -            & -            \\ 
\rowcolor{colorSnd}HRNet + T.      \cite{sun2019deep}                            & CVPR'19                                    & HRNet-W48                                        & 384$\times$384           &  94.46            &  63.60        & 60.69                         & 64.12                            & 95.90              &      0.93                    \\
\rowcolor{colorFst} RSB-Pose-HRNet                              &                                            & HRNet-W48                                        & 384$\times$384                  &   181.80                      &   72.77                          & \underline{28.14} \scriptsize{\textcolor{red}{32.55$\downarrow$}}             &  \underline{32.92}  \scriptsize{\textcolor{red}{31.20$\downarrow$}}            &  \textbf{98.09} \scriptsize{\textcolor{red}{2.19$\uparrow$}}            &   \textbf{0.99} \scriptsize{\textcolor{red}{0.06$\uparrow$}}           \\
\rowcolor{colorSnd}ResNet + T.  \cite{xiao2018simple}                               & ECCV'18                                    & ResNet-152                                       & 384$\times$384                   & 102.12                   & 68.64                       & 58.18         & 59.49         & 95.95        & 0.88         \\
\rowcolor{colorFst}  RSB-Pose-ResNet152                          &                                            & ResNet-152                                       & 384$\times$384                   & 190.83                   & 77.88                       & \textbf{27.00} \scriptsize{\textcolor{red}{31.18$\downarrow$}}         & \textbf{29.33} \scriptsize{\textcolor{red}{30.16$\downarrow$}}          & \underline{97.40} \scriptsize{\textcolor{red}{1.45$\uparrow$}}        & \underline{0.98} \scriptsize{\textcolor{red}{0.10$\uparrow$}}     \\
\Xhline{0.12em}
\end{tabular}}
\end{table*}
\linespread{1}

\subsubsection{Implementation Details}
\noindent Our RSB-Pose method is implemented by PyTorch \cite{paszke2017automatic}. We conduct experiments with two 2D backbones: ResNet \cite{he2016deep} and HRNet \cite{sun2019deep}, each at two different input scales. We refer to it as ``RSB-Pose-\textsc{2D Backbone}'' in tables. ResNet models are pre-trained like \cite{iskakov2019learnable}, while HRNet models are pre-trained in MPII \cite{andriluka14cvpr} and H36M. V2V \cite{moon2018v2v} is employed as the 3D convolution network in the SCE module. First, the PT is pre-trained with combined training sets of MHAD and H36M. This combination ensures a boarder coverage of 3D poses. The pre-training process lasts for 200 epochs and employs the AdamW optimizer \cite{loshchilov2017decoupled} with a learning rate $10^{-3}$. Subsequently, the training of the whole framework involves two distinct steps: pre-training of the SCE module and end-to-end training of the entire network. The SCE module is trained combined with 2D backbone and Triangulation components. The training undergoes with the supervision of Eq. \ref{eq_preloss} using Adam Optimizer with a learning rate $10^{-4}$ for 2D backbone and $10^{-3}$ for others. Reload all pre-trained weights, the whole framework is finally trained end-to-end. It is conducted under the $10^{-4}$ learning rate. 
It should be noted that when the test dataset changes, the above framework training process is carried out separately on different training datasets. Specifically, we trained on MHAD for 10 epochs and on H36M for 6 epochs.

\subsection{Quantitative Evaluation}

\subsubsection{Results on the MHAD Dataset}
\noindent Due to the limited number of methods specifically addressing binocular 3D HPE, we select several state-of-the-art (SOTA) multiview methods for comparison. To ensure a fair comparison, these methods are fine-tuned in MHAD, training details can be found in Sec. \ref{sec:appendix_training_detail} of Appendix.

As depicted in Tab. \ref{tb:compare_with_sota_mhad}, our RSB-Pose achieves the highest accuracy in both 3D and 2D pose estimation, regardless of input size. 
Notably, RSB-Pose-ResNet50, with $256\times 256$ input images, even surpasses the best-performing Volume-T. whose input is $384\times 384$. For the four metrics used to measure effectiveness, our RSB-Pose significantly outperforms the corresponding baselines, as highlighted in red. This demonstrates the role of RSB-Pose in improving 3D pose estimation and enhancing 2D pose view consistency regardless of the 2D backbones. Regarding efficiency metrics, at a resolution of 256, RSB-Pose requires a relatively higher amount of MACs and Params. However, this resource demand remains within an acceptable range. At a resolution of 384, RSB-Pose consumes fewer computing resources than Volume-T due to the concentrated disparity range, while achieving higher 3D pose accuracy.

Notably, the performance of SOTA methods here is significantly degraded compared to the 4-view performance in their papers. Upon closer analysis, it appears that there is a notable probability that the human area fusion module of TPPT may filter out a portion of the human body. The most likely regions to be filtered out are the limb joints, which can lead to poor JDR performance in estimating keypoints such as the elbow, wrist, nose, and head (see Tab. \ref{tb:compare_with_sota_mhad_detail} in Appendix). The fusion approach employed by Epipolar-T. and AdaFuse involves enhancing features or heatmaps by considering another view under geometric constraints. This fusion strategy is highly effective in multiview scenarios where multiple views provide abundant information to help filter out incorrect estimations. However, in the short-baseline binocular environment, two similar views may lead to a further increase in the 2D errors, which in turn results in significant 3D estimation errors. Our SCE module goes beyond merely enhancing features. It facilitates a collaborative decision-making process between the features from both views, allowing for a more comprehensive utilization of binocular visual and geometric information. 

\linespread{1.4}
\begin{table*}[t]
\caption{\textbf{Quantitative Comparison with SOTA methods on H36M datasets.} The best results within the same scale are highlighted in \textbf{bolded}, and the second best is \underline{underlined}. Our baseline and RSB-Pose are emphasized by \sethlcolor{colorSnd}\hl{light green} and \sethlcolor{colorFst}\hl{green}, respectively.}
\label{tb:compare_with_sota_h36m}
\centering
\resizebox{\linewidth}{!}{
\begin{tabular}{l|l|l|c|cc|cc|cc}
\Xhline{0.12em}
\multicolumn{1}{c|}{\multirow{2}{*}{Method}} & \multicolumn{1}{c|}{\multirow{2}{*}{Venue}} & \multicolumn{1}{c|}{\multirow{2}{*}{2D Backbone}} & \multirow{2}{*}{Input Size} & \multirow{2}{*}{MACs (G)} & \multirow{2}{*}{Params (M)} & \multicolumn{2}{c|}{3D Pose}   & \multicolumn{2}{c}{2D Pose} \\
\multicolumn{1}{c|}{}                        & \multicolumn{1}{c|}{}                       & \multicolumn{1}{c|}{}                             &                             &                          &                             & MPJPE\_re (mm) $\downarrow$ & MPJPE\_ab (mm) $\downarrow$& JDR (\%) $\uparrow$     & SIM\_cos  $\uparrow$    \\
\hline
TPPT  \cite{ma_ppt_2022}                                       & ECCV'22                                    & TokenPose                                        & 256$\times$256               & 9.70                     & 7.75                        & 42.10         & 40.72         & 93.32        & 0.87         \\
Epipolar-T.    \cite{he_epipolar_2020}                              & CVPR'20                                    & ResNet-50                                        & 256$\times$256                    & 52.28                    & 34.07                       & 37.97         & 41.22         & \textbf{96.40}        & 0.88         \\
\rowcolor{colorSnd}HRNet + T.   \cite{sun2019deep}                               & CVPR'19                                    & HRNet-W32                                        & 256$\times$256                     & 20.68                    & 28.55                       &   38.61            &    40.15           &   94.30           &  0.89            \\
\rowcolor{colorFst} RSB-Pose-HRNet                              &                                            & HRNet-W32                                        & 256$\times$256                     &  105.26                        &   37.70                          &  \underline{33.16} \scriptsize{\textcolor{red}{5.45$\downarrow$}}            & \underline{36.84} \scriptsize{\textcolor{red}{3.31$\downarrow$}}              &     \underline{96.18} \scriptsize{\textcolor{red}{1.88$\uparrow$}}         &  \textbf{0.94}  \scriptsize{\textcolor{red}{0.05$\uparrow$}}           \\
\rowcolor{colorSnd}ResNet + T.   \cite{xiao2018simple}                               & ECCV'18                                    & ResNet-50                                        &  256$\times$256                      & 25.87                    & 33.99                            &  36.52             &   37.04      &94.07     & 0.82         \\
\rowcolor{colorFst} RSB-Pose-ResNet50                          &                                            & ResNet-50                                        & 256$\times$256                & 111.45                   & 43.24                       & \textbf{32.80}  \scriptsize{\textcolor{red}{3.72$\downarrow$}}        & \textbf{35.01} \scriptsize{\textcolor{red}{2.03$\downarrow$}}         & 94.82  \scriptsize{\textcolor{red}{0.75$\uparrow$}}       & \underline{0.92} \scriptsize{\textcolor{red}{0.10$\uparrow$}}         \\ \hline
AdaFuse    \cite{zhang_adafuse_2021}                                 & IJCV'20                                    & ResNet-152                                       & 384$\times$384                  & 149.48                   & 69.66                       & \underline{28.52}         & \underline{30.27}         & 94.25        & 0.89         \\
Algebraic-T.   \cite{iskakov2019learnable}                             & ICCV'19                                    & ResNet-152                                       & 384$\times$384                 & 104.92                   & 79.52                       & 29.44         & 31.24         & \underline{95.81}        & 0.79         \\
Volume-T.    \cite{iskakov2019learnable}                               & ICCV'19                                    & ResNet-152                                       & 384$\times$384                   & 257.15                   & 80.59                       & 30.97         & 32.63         & -            & -            \\
\rowcolor{colorSnd}HRNet + T.      \cite{sun2019deep}                            & CVPR'19                                    & HRNet-W48                                        & 384$\times$384                  &     94.46                     &  63.60                           &   31.97            &   33.28            &  93.26            &     0.89         \\
\rowcolor{colorFst} RSB-Pose-HRNet                              &                                            & HRNet-W48                                        & 384$\times$384                  &       331.56                   &      72.77                       &     29.50  \scriptsize{\textcolor{red}{2.47$\downarrow$}}         & 30.73 \scriptsize{\textcolor{red}{2.55$\downarrow$}}              &  95.54  \scriptsize{\textcolor{red}{2.28$\uparrow$}}           &  \textbf{0.95} \scriptsize{\textcolor{red}{0.06$\uparrow$}}            \\
\rowcolor{colorSnd}ResNet + T.  \cite{xiao2018simple}                               & ECCV'18                                    & ResNet-152                                       & 384$\times$384                   & 102.12                   & 68.64                       & 29.61         & 31.51         & \underline{95.81}        & 0.79         \\
\rowcolor{colorFst}  RSB-Pose-ResNet152                          &                                            & ResNet-152                                       & 384$\times$384                   & 339.24                   & 77.87                       & \textbf{27.89} \scriptsize{\textcolor{red}{1.72$\downarrow$}}         & \textbf{30.07} \scriptsize{\textcolor{red}{1.44$\downarrow$}}         & \textbf{95.93} \scriptsize{\textcolor{red}{0.12$\uparrow$}}        & \underline{0.94}  \scriptsize{\textcolor{red}{0.15$\uparrow$}}      \\
\Xhline{0.12em}
\end{tabular}
}
\end{table*}
\linespread{1}

\subsubsection{Results on the H36M Dataset}
\noindent In our experiments, we utilize the models of SOTA methods with their official weights, making the sole adjustment of changing the camera pair to the binocular settings for evaluation. Tab. \ref{tb:compare_with_sota_h36m} shows the quantitative comparison of our RSB-Pose and other SOTA methods. 
Under the input scale of 256, our RSB-Pose-ResNet50 outperforms TPPT \cite{ma_ppt_2022} and Epipotar-T. \cite{he_epipolar_2020} by more than $5mm$ regardless MPJPE\_re or MPJPE\_ab. 
Regarding the JDR metric, the performance is less favorable. This discrepancy attributes to the variations of the groundtruth bounding box in different methods, leading to the differences in the ratio of the human subject within the bounding box. But in SIM\_cos, the superiority of our method is significant. Under the input scale of 384, which captures more fine-grained features, our RSB-Pose-ResNet152 achieves the best performance in MPJPE\_re, outperforming other methods by a margin of at least $2.2\%$. Moreover, when evaluated under the same bounding box setting, RSB-Pose consistently exceeds other methods in JDR. However, it cannot be ignored that the MACs efficiency of RSB-Pose in H36M is suboptimal, which will be discussed in detail in Sec. \ref{sec:discussion}.

\subsubsection{Robustness Analysis}
A comparative analysis between Tab. \ref{tb:compare_with_sota_mhad} and Tab. \ref{tb:compare_with_sota_h36m} reveals notable insights. The 3D pose accuracy of RSB-Pose in MHAD (with a $200mm$ baseline) remains consistent with that in H36M (with a $3000mm$ baseline), while most other SOTA methods suffer a drop in accuracy with the shorter baseline. Specifically, the MPJPE\_ab of RSB-Pose-ResNet152 is $29.33mm$ and $30.07mm$ in MHAD and H36M, respectively. The MPJPE\_ab of Algebraic-T. is $31.24mm$ in H36M, but increases to $51.69mm$ in MHAD. This observation demonstrates that our method is more robust than other SOTA methods in scenarios of short-baseline setups.

Volume-T. exhibits robustness comparable to RSB-Pose. We attempt to further compare these two methods theoretically. There are some similarities between Volume-T. and our approach in feature construction. Both methods involve a volume feature, allowing the 3D point to be determined by all views simultaneously, which enhances robustness. However, there are significant differences as well. In Volume-T., the volume feature is created by projecting the image features back into 3D space. In contrast, our approach constructs the SVF in the disparity space. The advantages of using the disparity space include: 1) unlike the interpolation is needed when projecting to 3D space, the correspondence of two views is more flexible in the disparity space; 2) the SVF is built by sift-window which is much easier; 3) an initial volume center is no more needed; 4) volume size is reduced in short-baseline settings due to the focus range of disparity, consequently the MACs are reduced, as shown in Tab. \ref{tb:compare_with_sota_mhad}. These benefits make our method more efficient.

\begin{figure*}[!t]
    \centering
    \includegraphics[width=\linewidth]{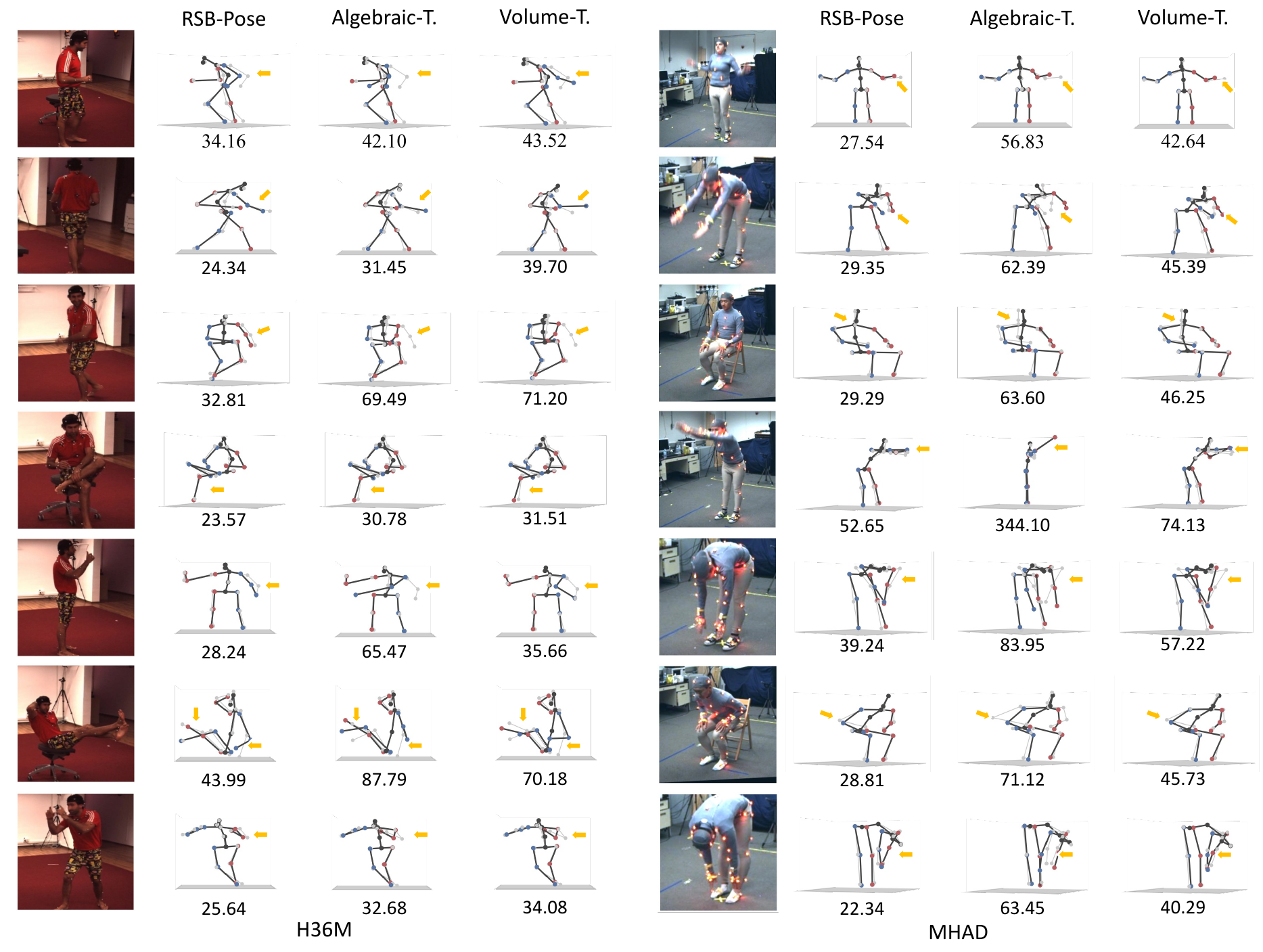}
    \vspace{-1.5em}
    \caption{Qualitative comparison with SOTA methods. The images are all captured from the left view. The number under each pose corresponds to the MPJPE\_re result. The \textcolor{White}{\sethlcolor{mygray}\hl{gray}} skeleton represents the groundtruth, while the \sethlcolor{mygray}\hl{black} skeleton represents the estimated pose. In the black skeleton, right joints are marked in \sethlcolor{myred}\hl{red}, and left joints are marked in \sethlcolor{myblue}\hl{blue}. The left half shows results on the H36M dataset and the right half is on the MHAD dataset. \sethlcolor{yellow}\hl{Yellow} arrows indicate parts of significant improvement in our method compared to SOTA methods.}
    \label{fig_compare_with_sota}
\end{figure*}

\subsection{Qualitative Evaluation}
\noindent To intuitively compare the results with SOTA methods, we further visualize some 3D poses generated from our RSB-Pose, Algebraic-T., and Volume-T. \cite{iskakov2019learnable} respectively. As shown in Fig. \ref{fig_compare_with_sota}, the left samples are from the H36M dataset, while the right samples are from MHAD. In general, RSB-Pose excels in estimating the limb joints, which are the most flexible, including the elbow, wrist, knee, ankle, and head. Even in cases of heavy occlusion in both views, such as the $1^{st}$, $3^{th}$ and $6^{th}$ examples on H36M and the last four examples on MHAD, RSB-Pose provides superior and plausible results.

\subsection{Ablation Study} \label{sec:ablation_study}

\subsubsection{The impact of SCE module} \label{sec:impact_of_SCE}
\noindent We investigate the influence of two introduced modules respectively. To begin with, we establish the baseline methods, named Baseline-50 and Baseline-152, which differ in terms of the backbone used. The baseline framework comprises three modules: a 2D backbone ResNet50; a 2D keypoints regression head, essentially a $1 \times 1$ convolution layer; and a Triangulation part, to reconstruct 3D keypoints. Next, to evaluate the impact of the SCE module, we replace the 2D keypoints regression head with it. Additionally, we also explore two methods: one with the inclusion of an AM and the other without. The experimental results are presented in Tab. \ref{tb:impact_of_each_module} \sethlcolor{colorFst}\hl{green columns}. 

Remarkably, even in the absence of an AM, the SCE demonstrates substantial improvements in MPJPE\_ab across all keypoints, achieving enhancements of $24\%$ and $48\%$ for the respective backbone configurations. This performance trend holds consistently when evaluating the MPJPE\_re protocol, where improvements of $26\%$ and $52\%$ are observed, once again showcasing the effectiveness of the SCE module. Notably, these improvements are consistent for both unoccluded and occluded keypoints, demonstrating the robustness of the SVF feature generation approach, which concatenates binocular features. In another word, the enhancement of view consistency, including the constraint on Y-axis results for keypoints, proves to be a valuable addition to the model.

\begin{table}[t]
\caption{\textbf{The Impact of Each Module.} The impacts of SCE and PPT are emphasized by \sethlcolor{colorFst}\hl{green} and \sethlcolor{colorSnd}\hl{light green}. ``n\_occ" and ``occ" denote the unocclued keypoints and occluded ones.}
\label{tb:impact_of_each_module}
\centering
\setlength{\tabcolsep}{1pt}
\renewcommand{\arraystretch}{1.2}
\resizebox{0.45\textwidth}{!}{

\begin{tabular}{l|ccc|ccc}
\Xhline{0.12em}
\multicolumn{1}{l|}{\multirow{2}{*}{\textbf{MPJPE\_ab} (mm) $\downarrow$}} & \multicolumn{2}{c}{SCE} &\multirow{2}{*}{PPT} & \multirow{2}{*}{all} & \multirow{2}{*}{n\_occ} & \multirow{2}{*}{occ} \\
\multicolumn{1}{c|}{}                           & w/o AM    & w/ AM   &          &                      &                         &                      \\ \hline
Baseline-50                                    &             &           &            & 63.18                & 62.99                   & 73.13                \\ 
\multirow{3}{*}{RSB-Pose-ResNet50}                                 & \checkmark           &           &           & \fs47.61                & \fs47.42                   & \fs54.50                \\
                                               &            & \checkmark         &           & \fs35.40                & \fs 35.25                   & \fs 40.73                \\
                                               &             & \checkmark         & \checkmark          & \nd32.10                & \nd31.94                   & \nd37.78                \\ \hline
Baseline-152                                   &             &           &            & 59.00                & 58.79                   & 69.99                \\ 
\multirow{3}{*}{RSB-Pose-ResNet152}                                 & \checkmark           &           &            & \fs30.40                & \fs30.24                   & \fs36.13                \\
                                               &             & \checkmark         &            & \fs29.78                & \fs29.60                   & \rd36.19                \\
                                               &            & \checkmark         & \checkmark          & \nd29.33 & \nd29.17  & \rd35.06                    \\ \Xhline{0.1em}
\multicolumn{1}{l|}{\multirow{2}{*}{\textbf{MPJPE\_re} (mm) $\downarrow$}} & \multicolumn{2}{c}{SCE} & \multirow{2}{*}{PPT} & \multirow{2}{*}{all} & \multirow{2}{*}{notocc} & \multirow{2}{*}{occ} \\
\multicolumn{1}{c|}{}                           & w/o mask    & w/ mask   &         &                      &                         &                      \\  \hline
Baseline-50                                    &             &           &            & 62.70                & 62.34                   & 75.38                \\ 
\multirow{3}{*}{RSB-Pose-ResNet50}                                & \checkmark           &           &            & \fs46.16                & \fs46.05                   & \fs50.19                \\
                                               &             & \checkmark         &            & \fs31.10                & \fs30.93                   & \fs37.23                \\
                                               &             & \checkmark         & \checkmark          & \nd26.93                & \nd26.80                   & \nd31.42                \\ \hline
Baseline-152                                   &             &           &            & 58.18                & 49.93                   & 58.41               \\ 
\multirow{3}{*}{RSB-Pose-ResNet152}                                  & \checkmark           &         &            & \fs27.98                & \fs27.80                   & \fs34.59                \\
                                               &             & \checkmark         &            & \fs27.20               & \fs27.02                   & \rd33.92                \\
                                               &             & \checkmark         & \checkmark          & \nd27.00& \nd26.93  & \rd29.46     \\ \Xhline{0.12em}              
\end{tabular}}
\end{table}

The addition of the AM also leads to improvements in accuracy, although these improvements vary between different backbones. Particularly, the enhancement is significantly more pronounced in RSB-Pose-ResNet50 compared to RSB-Pose-ResNet152, with a reduction in MPJPE\_re of $15.05 mm$  versus $0.78 mm$. Our analysis indicates that the AM plays a crucial role in filtering out disruptive features, helping the SVF accurately regress results by reducing interference from similar background and foreground features. However ResNet-152 possesses a more powerful feature extraction capability compared to ResNet-50, and as a result, the confusion of features no longer exists.

\begin{figure}[!t]
    \centering
    \includegraphics[width=0.8\linewidth]{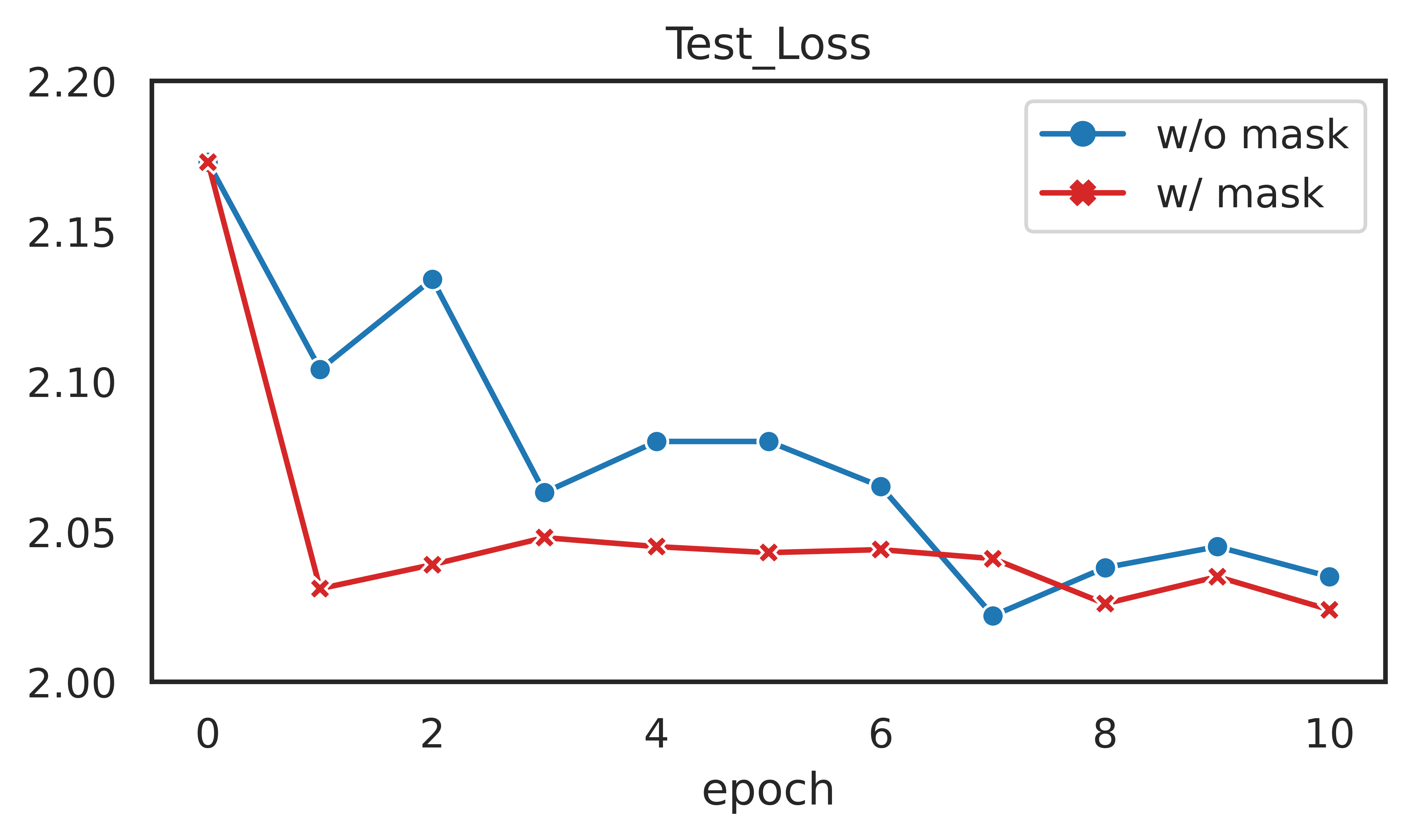}
    \vspace{-0.5em}
    \caption{Convergence trend of loss in the test dataset during the RSB-Pose-50 training process.}
    \label{fig_test_loss_during_learning}
\end{figure}

Additionally, AM plays a role in the convergence during training. As shown in Fig. \ref{fig_test_loss_during_learning}, the total loss convergence trend of the MHAD test dataset is more stable with the addition of the AM, which is different from the situation when there is no mask. The AM directs SVF to concentrate on the relevant body parts of interest, thus effectively facilitating the 3D regression. Without the use of masks, the target features become too sparse, making it difficult to achieve convergence.

\subsubsection{The impact of PPT module} \label{sec:impact_of_PT}
\noindent We first conduct ablation experiments (see Sec. \ref{sec:appendix_impact_of_PT_network} in Appendix) to design the network structure of PT through the pre-training task and then verify the effectiveness of PPT in 3D pose refinement. Based on the experiments in Sec. \ref{sec:impact_of_SCE}, we incorporate the PPT module to further refine the initial 3D poses, with end-to-end training for 10 epochs. As shown in Tab. \ref{tb:impact_of_each_module} \sethlcolor{colorSnd}\hl{light green parts}. In the case of RSB-Pose-ResNet50, the refinement process leads to enhanced accuracy for all keypoints, both unoccluded and occluded joints. This improvement is particularly remarkable, resulting in a $3.3 mm$ reduction in MPJPE\_ab and a $4.17 mm$ reduction in MPJPE\_re. For RSB-Pose-ResNet152, although the overall improvement in accuracy is not as significant, there is still an enhancement, with a $1.5\%$ MPJPE\_ab reduction and a $0.7\%$ MPJPE\_re reduction. Notably, the refinement process has a more pronounced effect on occluded keypoints in RSB-Pose-ResNet152, with accuracy improvements of $3.1\%$ in MPJPE\_ab and $13.1\%$ in MPJPE\_re, emphasized by \sethlcolor{colorTrd}\hl{yellow}. The observed improvements suggest that the PPT module is effective in promoting the overall pose quality and it is particularly valuable for occluded keypoints.

\begin{figure}[!t]
    \centering
    \includegraphics[width=0.8\linewidth]{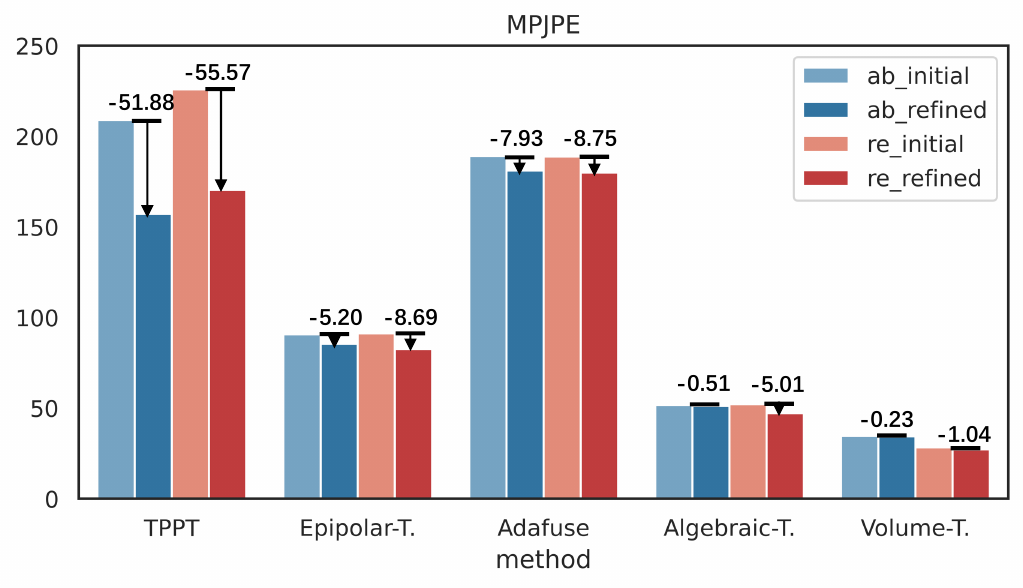}
    \vspace{-0.5em}
    \caption{The impact of PPT refinement in MHAD. The \sethlcolor{myblue}\hl{blue} histograms represent MPJPE\_ab and the \sethlcolor{myred}\hl{red} ones denote MPJPE\_re. The \textcolor{White}{\sethlcolor{mygray}\hl{light-colored}} bars indicate the initial 3D pose, while the \sethlcolor{mygray}\hl{dark-colored} bars represent the refined poses. The value indicates the MPJPE enhancement after refinement.}
    \label{fig_validation_of_pt_refinement}
\end{figure}

To further validate the generalization of the PPT module, we conduct the refinement on five existing methods. PPT is directly applied to refine the 3D poses generated by these methods without training. Fig. \ref{fig_validation_of_pt_refinement} illustrates the differences between the initial poses and the refined poses in the MHAD dataset. The PPT leads to performance enhancements across all methods. We get the same conclusion in the MHAD\_occ dataset, which can be found in Sec. \ref{sec:appendix_ppt_val} of Appendix. In summary, these results demonstrate the versatility of the PPT module as a plug-and-play component and its ability to effectively enhance overall pose coherence.

\begin{figure}[!t]
    \centering
    \includegraphics[width=0.45\textwidth]{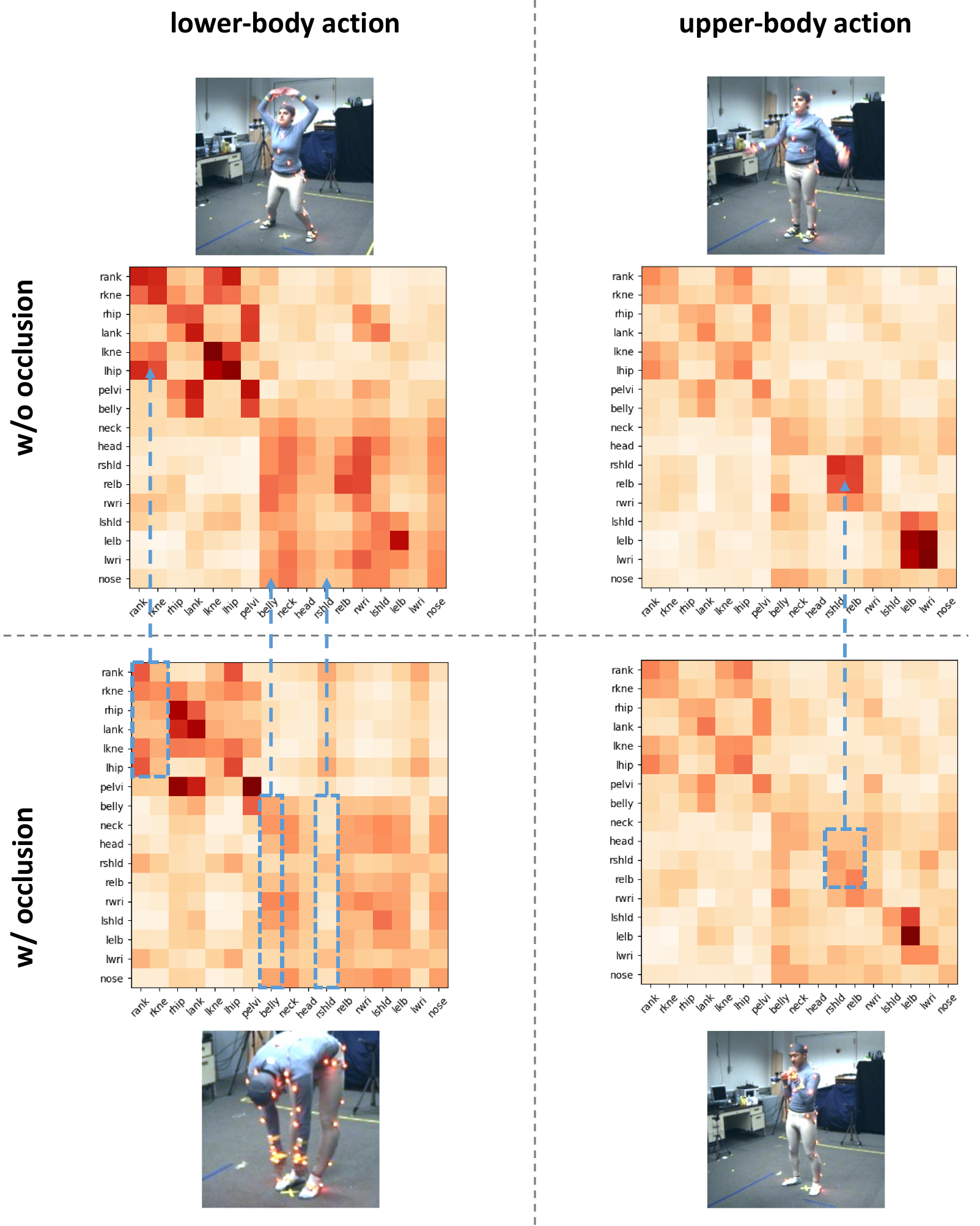}
    \vspace{-0.5em}
    \caption{Differences of attention map between occluded and unoccluded situations. The attention map denotes the spatial dependency from row keypoints to col keypoints. The first row represents the situation where all keypoints are visible. The second row represents the situation where some keypoints are occluded. The \sethlcolor{myblue}\hl{blue} dashed box in the second row indicates that occluded joints receive reduced attention compared to their unoccluded cases.}
    \label{fig_attentionmap_withocc}
\end{figure}

However, the underlying principle of refinement through the PPT remains somewhat ambiguous, and it is uncertain whether correlations between joints can be effectively perceived. Hence, we conduct the qualitative experiments to explore it. We visualize the sum of the multi-head attention maps in the first layer of PT, shown in Fig. \ref{fig_attentionmap_withocc}. In the first row, representing the situation that all keypoints are visible, two main types of attention patterns are observed. The first type focuses on lower-body actions, with greater associations among leg joints. The second type highlights upper-body movements, emphasizing stronger associations between arm joints. Consequently, PPT demonstrates its capability to describe various actions. We also observe a commonality across both types: a stronger correlation between neighboring joints, such as the elbow-wrist and ankle-knee pairs. This indicates that PPT prioritizes relationships between adjacent joints, regardless of body region or action type. Moving on to the second row, which represents situations involving occluded joints, a notable observation is that other joints tend to pay less attention to the occluded keypoints. For instance, as shown in the second case, where the right shoulder and elbow are occluded, there is an obvious reduction in attention towards these joints when compared to the visible case in the first row. In summary, these results provide evidence that the PPT module can establish meaningful correlations between joints based on the anatomical skeleton, and it then effectively filters out the occluded joints.

\subsubsection{The impact of the pre-training strategy}\label{sec:effect_of_pretrain}
\noindent To investigate the influence of iterative masking times $T$ during pre-training, we conduct experiments with varying masking times from 1 to 5. All experiments are trained using a combined dataset of H36M and MHAD and subsequently tested on the MHAD test dataset. The input 3D pose is masked at a ratio of $0.4$, which is determined based on experiments in Appendix Sec. \ref{sec:appendix_pretraining}. The results produced are listed in Tab. \ref{tb:impact_of_iterative_mask_time}. $T=1$ refers to the situation only with initial masking. The role of iterative masking is evident, as the performance continues to be improved when additional masking iterations are applied, regardless of the specific number of repetitions. This demonstrates the effectiveness of iterative masking in pre-training, acting as a data augmentation. An intriguing finding is that increasing the iterative time does not necessarily lead to better results. The performance peaks at $T=2$, indicating that applying the mask twice is the optimal strategy. Our analysis suggests that by stacking more iterations of masking, a moment will be reached where no more nodes need to be masked. In other words, the training dataset is no longer enlarged, which may lead to overfitting of the error distribution during training, thus affecting the performance when applied to the test data.

\begin{table}[t]
\caption{\textbf{The Impact of Iterative Masking.} ``Iter\_T" denotes repeat times $T$ and $Iter\_T = 1$ refers to the condition involves only the initial masking. The experiments are trained on H36M and MHAD training datas and tested on the MHAD testing datas.}
\label{tb:impact_of_iterative_mask_time}
\centering
\setlength{\tabcolsep}{5pt}
\renewcommand{\arraystretch}{1.1}
\resizebox{0.45\textwidth}{!}{
\begin{tabular}{l|ccccc}
\Xhline{0.12em}
Iter\_T (Mask\_Ratio=0.4)       & 1     & 2     & 3     & 4     & 5     \\ \hline
MPJPE\_ab (mm) $\downarrow$& 29.41 & 20.19 & 23.86 & 25.78 & 28.29 \\
MPJPE\_re (mm) $\downarrow$& 28.81 & 20.12 & 24.01 & 24.89 & 29.49 \\\Xhline{0.12em}
\end{tabular}}
\end{table}
\begin{table}[t]
\caption{\textbf{The Impact of Cascade Strategy.} ``all" and ``occ" denote MHAD and MHAD\_occ. The masking cascade strategy is emphasized by \sethlcolor{colorFst}\hl{green}, while the others are \sethlcolor{colorSnd}\hl{light green}.}
\label{tb:impact_of_combination_strategy}
\centering
\setlength{\tabcolsep}{1pt}
\renewcommand{\arraystretch}{1.2}
\resizebox{0.45\textwidth}{!}{
\begin{tabular}{l|cc|cc|cc}
\Xhline{0.12em}
\multicolumn{1}{l|}{\multirow{2}{*}{Method}} & \multirow{2}{*}{pretrained} & \multirow{2}{*}{relative} & \multicolumn{2}{c|}{MPJPE\_ab (mm) $\downarrow$} & \multicolumn{2}{c}{MPJPE\_re (mm) $\downarrow$} \\
\multicolumn{1}{c|}{}                        &                             &                           & all            & occ           & all           & occ           \\\hline
                                             &                             &                           & 38.95          & 43.97         & 36.81         & 43.41         \\
occ\_mask                                    & \checkmark                           &                           & \fs36.91          & \fs47.43         & \fs34.33         & \fs42.66         \\
conf\_mask                                   & \checkmark                           &                          & \fs39.55          & \fs42.10         & \fs34.66         &\fs 38.45         \\
no\_mask                                     & \checkmark                           &                \          &\fs 36.18          & \fs39.20         & \fs32.58         & \fs36.20         \\
no\_mask                                     & \checkmark                           & \checkmark                         & \nd33.90          & \nd37.05         & \nd32.59         & \nd36.41         \\
no\_mask                                     &                            &  \checkmark                         & \nd46.54          & \nd46.13         & \nd39.78         & \nd41.44    \\ \Xhline{0.12em}    
\end{tabular}}
\end{table}

\subsubsection{The impact of the cascade strategy of PPT}
\noindent The primary objective of the PPT module is to improve the accuracy of occluded joints, which are prone to poor estimation due to the absence of visual information. To achieve this goal, we aim to inject pose coherence, which is ignored by the Triangulation. Intuitively, during inference, one might expect that the occluded joints of the initial 3D pose should be masked and then processed by the PPT, mirroring the pre-training process. However, our experiments yield unexpected results. 

We conduct three cascade strategies: occ\_mask, conf\_mask and no\_mask. In occ\_mask, we mask the occluded keypoints, while in no\_mask, we use the whole initial 3D pose as-is. Recognizing that poor estimation results can have various underlying causes besides occlusion, we introduced the conf\_mask strategy. Here, the confidence of each keypoint is generated by a regression head, and then integrated into the attention map in the first layer of the PT module, reducing reliance on less trustworthy keypoints. As depicted in Tab. \ref{tb:impact_of_combination_strategy}, the first row presents results from RSB-Pose-50 without refinement, while \sethlcolor{colorFst}\hl{the subsequent rows} show performance after refinement using different strategies. In the MHAD dataset, the conf\_mask strategy yields worse results in MPJPE\_ab, likely due to the negative impact of unreliable confidence scores. In the MHAD\_occ, the occ\_mask strategy achieves the poorest results, indicating that an occlusion-based masking strategy hampers final performance due to the complete absence of visual information. Surprisingly, the no\_mask strategy outperforms the others by a significant margin in both two datasets. We attribute this improvement to the retention of visual information included in the initial estimate. Furthermore, as demonstrated in Section \ref{sec:impact_of_PT}, the PPT module is capable of identifying occluded joints. Consequently, it appears that masking is no longer necessary during inference.

Moreover, we change the 3D pose from absolute to \sethlcolor{colorSnd}\hl{relative} in pre-training and inference simultaneously. As expected, the relative format surpasses the absolute one with over $2 mm$ MPJPE\_ab, because the pose space becomes more constrained to promote the exploration of joint correlations. Finally, we incorporate the PT module without pre-training into the entire framework and conduct end-to-end training for 10 epochs. As expected, a decrease is resulted in accuracy. We theorize that during pre-training, PT understands the joint correlations of human anatomy skeleton, and during end-to-end refinement, it focuses on learning the specific error distribution of the initial 3D pose. However, without pre-training, PT lacks the pre-learned pose prior, which, in turn, impedes the parameter optimization process. This underscores the significance of the joint correlations learning through pre-training.

\subsection{Discussions}\label{sec:discussion}
\noindent It is important to note that our RSB-Pose approach has several limitations and deserves further improvement. Firstly, our method is more suitable for short-baseline binocular setups than for wider-baselines. As the baseline increases, resulting in a larger angle between the cameras' optical shafts, challenges arise in terms of rectification and expanding the required disparity range. For example, the MHAD dataset has a disparity range of 17 pixels, while H36M exhibits a much larger range of 60 pixels. The $D$ dimension in SVF is decided by the disparity range. Hence the extended disparity range consumes more computational resources during the regression of SVF. As shown in Tab. \ref{tb:compare_with_sota_mhad} and Tab. \ref{tb:compare_with_sota_h36m}, the MACs of RSB-Pose in H36M are higher than in MHAD, although the parameters remain the same. Moreover, it is crucial to acknowledge that RSB-Pose is not compatible with multiview settings, as our SVF generation relies on rectification, which enforces Y-axis constraints for binocular corresponding keypoints. Therefore, expanding our method to multiview settings is a key purpose for future refinement, making it more generalizable.

\section{Conclusion}
\noindent We present RSB-Pose, a specialized method designed for short-baseline binocular 3D HPE. Our approach incorporates an SCE module to robust 3D results by generating view-consistent 2D binocular keypoints. Within the SCE module, the disparity is utilized to represent two-view 2D correspondences, and the SVF is introduced to concatenate binocular features under various disparities and finally regress the binocular 2D co-keypoints. Additionally, we introduce a PPT to refine 3D poses by injecting pose coherence perception, rendering them robust against occlusions. We evaluate RSB-Pose on two benchmark datasets: H36M and MHAD, and conduct extensive experiments to demonstrate its effectiveness and occlusion handling capability. Our findings prove the efficacy of SCE in facilitating 3D keypoint reconstruction, while demonstrating that the PPT is capable of modeling joint correlations in a meaningful manner.


{\small
\bibliographystyle{IEEEtran}
\bibliography{IEEEabrv, RSB-POSE/refs}

\begin{thebibliography}{10}
\providecommand{\url}[1]{#1}
\csname url@samestyle\endcsname
\providecommand{\newblock}{\relax}
\providecommand{\bibinfo}[2]{#2}
\providecommand{\BIBentrySTDinterwordspacing}{\spaceskip=0pt\relax}
\providecommand{\BIBentryALTinterwordstretchfactor}{4}
\providecommand{\BIBentryALTinterwordspacing}{\spaceskip=\fontdimen2\font plus
\BIBentryALTinterwordstretchfactor\fontdimen3\font minus
  \fontdimen4\font\relax}
\providecommand{\BIBforeignlanguage}[2]{{%
\expandafter\ifx\csname l@#1\endcsname\relax
\typeout{** WARNING: IEEEtran.bst: No hyphenation pattern has been}%
\typeout{** loaded for the language `#1'. Using the pattern for}%
\typeout{** the default language instead.}%
\else
\language=\csname l@#1\endcsname
\fi
#2}}
\providecommand{\BIBdecl}{\relax}
\BIBdecl

\bibitem{WANG2021103225}
J.~Wang, S.~Tan, X.~Zhen, S.~Xu, F.~Zheng, Z.~He, and L.~Shao, ``Deep 3d human
  pose estimation: A review,'' \emph{Computer Vision and Image Understanding},
  vol. 210, p. 103225, 2021.

\bibitem{pavlakos2017coarse}
G.~Pavlakos, X.~Zhou, K.~G. Derpanis, and K.~Daniilidis, ``Coarse-to-fine
  volumetric prediction for single-image 3d human pose,'' in \emph{Proceedings
  of the IEEE conference on computer vision and pattern recognition}, 2017, pp.
  7025--7034.

\bibitem{martinez2017simple}
J.~Martinez, R.~Hossain, J.~Romero, and J.~J. Little, ``A simple yet effective
  baseline for 3d human pose estimation,'' in \emph{Proceedings of the IEEE
  international conference on computer vision}, 2017, pp. 2640--2649.

\bibitem{wu2021limb}
L.~Wu, Z.~Yu, Y.~Liu, and Q.~Liu, ``Limb pose aware networks for monocular 3d
  pose estimation,'' \emph{IEEE Transactions on Image Processing}, vol.~31, pp.
  906--917, 2021.

\bibitem{iskakov2019learnable}
K.~Iskakov, E.~Burkov, V.~Lempitsky, and Y.~Malkov, ``Learnable triangulation
  of human pose,'' in \emph{Proceedings of the IEEE/CVF international
  conference on computer vision}, 2019, pp. 7718--7727.

\bibitem{he_epipolar_2020}
Y.~He, R.~Yan, K.~Fragkiadaki, and S.-I. Yu, ``Epipolar transformers,'' in
  \emph{Proceedings of the ieee/cvf conference on computer vision and pattern
  recognition}, 2020, pp. 7779--7788.

\bibitem{zhang_adafuse_2021}
Z.~Zhang, C.~Wang, W.~Qiu, W.~Qin, and W.~Zeng, ``Adafuse: Adaptive multiview
  fusion for accurate human pose estimation in the wild,'' \emph{International
  Journal of Computer Vision}, vol. 129, pp. 703--718, 2021.

\bibitem{hartley_zisserman_2004}
R.~Hartley and A.~Zisserman, \emph{Multiple View Geometry in Computer Vision},
  2nd~ed.\hskip 1em plus 0.5em minus 0.4em\relax Cambridge University Press,
  2004.

\bibitem{qiu_cross_2019}
H.~Qiu, C.~Wang, J.~Wang, N.~Wang, and W.~Zeng, ``Cross view fusion for 3d
  human pose estimation,'' in \emph{Proceedings of the IEEE/CVF international
  conference on computer vision}, 2019, pp. 4342--4351.

\bibitem{remelli2020lightweight}
E.~Remelli, S.~Han, S.~Honari, P.~Fua, and R.~Wang, ``Lightweight multi-view 3d
  pose estimation through camera-disentangled representation,'' in
  \emph{Proceedings of the IEEE/CVF conference on computer vision and pattern
  recognition}, 2020, pp. 6040--6049.

\bibitem{ma_ppt_2022}
H.~Ma, Z.~Wang, Y.~Chen, D.~Kong, L.~Chen, X.~Liu, X.~Yan, H.~Tang, and X.~Xie,
  ``{PPT}: Token-pruned pose transformer for monocular and multi-view human
  pose estimation,'' in \emph{Computer Vision – {ECCV} 2022}, ser. Lecture
  Notes in Computer Science, S.~Avidan, G.~Brostow, M.~Cissé, G.~M. Farinella,
  and T.~Hassner, Eds.\hskip 1em plus 0.5em minus 0.4em\relax Springer Nature
  Switzerland, pp. 424--442.

\bibitem{zhang_mixste_2022}
J.~Zhang, Z.~Tu, J.~Yang, Y.~Chen, and J.~Yuan, ``Mixste: Seq2seq mixed
  spatio-temporal encoder for 3d human pose estimation in video,'' in
  \emph{Proceedings of the IEEE/CVF conference on computer vision and pattern
  recognition}, 2022, pp. 13\,232--13\,242.

\bibitem{tang_3d_nodate}
Z.~Tang, Z.~Qiu, Y.~Hao, R.~Hong, and T.~Yao, ``3d human pose estimation with
  spatio-temporal criss-cross attention,'' in \emph{Proceedings of the IEEE/CVF
  Conference on Computer Vision and Pattern Recognition}, 2023, pp. 4790--4799.

\bibitem{zhao_poseformerv2_2023}
Q.~Zhao, C.~Zheng, M.~Liu, P.~Wang, and C.~Chen, ``Poseformerv2: Exploring
  frequency domain for efficient and robust 3d human pose estimation,'' in
  \emph{Proceedings of the IEEE/CVF Conference on Computer Vision and Pattern
  Recognition}, 2023, pp. 8877--8886.

\bibitem{xue2022boosting}
Y.~Xue, J.~Chen, X.~Gu, H.~Ma, and H.~Ma, ``Boosting monocular 3d human pose
  estimation with part aware attention,'' \emph{IEEE Transactions on Image
  Processing}, vol.~31, pp. 4278--4291, 2022.

\bibitem{devlin2018bert}
J.~Devlin, M.-W. Chang, K.~Lee, and K.~Toutanova, ``Bert: Pre-training of deep
  bidirectional transformers for language understanding,'' \emph{arXiv preprint
  arXiv:1810.04805}, 2018.

\bibitem{tome2017lifting}
D.~Tome, C.~Russell, and L.~Agapito, ``Lifting from the deep: Convolutional 3d
  pose estimation from a single image,'' in \emph{Proceedings of the IEEE
  conference on computer vision and pattern recognition}, 2017, pp. 2500--2509.

\bibitem{zhou2017towards}
X.~Zhou, Q.~Huang, X.~Sun, X.~Xue, and Y.~Wei, ``Towards 3d human pose
  estimation in the wild: a weakly-supervised approach,'' in \emph{Proceedings
  of the IEEE international conference on computer vision}, 2017, pp. 398--407.

\bibitem{tekin2016structured}
B.~Tekin, I.~Katircioglu, M.~Salzmann, V.~Lepetit, and P.~Fua, ``Structured
  prediction of 3d human pose with deep neural networks,'' \emph{arXiv preprint
  arXiv:1605.05180}, 2016.

\bibitem{tekin2016direct}
B.~Tekin, A.~Rozantsev, V.~Lepetit, and P.~Fua, ``Direct prediction of 3d body
  poses from motion compensated sequences,'' in \emph{Proceedings of the IEEE
  Conference on Computer Vision and Pattern Recognition}, 2016, pp. 991--1000.

\bibitem{tu2023consistent}
Z.~Tu, Z.~Huang, Y.~Chen, D.~Kang, L.~Bao, B.~Yang, and J.~Yuan, ``Consistent
  3d hand reconstruction in video via self-supervised learning,'' \emph{IEEE
  Transactions on Pattern Analysis and Machine Intelligence}, vol.~45, no.~8,
  pp. 9469--9485, 2023.

\bibitem{kocabas2019self}
M.~Kocabas, S.~Karagoz, and E.~Akbas, ``Self-supervised learning of 3d human
  pose using multi-view geometry,'' in \emph{Proceedings of the IEEE/CVF
  conference on computer vision and pattern recognition}, 2019, pp. 1077--1086.

\bibitem{chen2019weakly}
X.~Chen, K.-Y. Lin, W.~Liu, C.~Qian, and L.~Lin, ``Weakly-supervised discovery
  of geometry-aware representation for 3d human pose estimation,'' in
  \emph{Proceedings of the IEEE/CVF conference on computer vision and pattern
  recognition}, 2019, pp. 10\,895--10\,904.

\bibitem{Kocabas_2020_CVPR}
M.~Kocabas, N.~Athanasiou, and M.~J. Black, ``Vibe: Video inference for human
  body pose and shape estimation,'' in \emph{CVPR}, June 2020.

\bibitem{zeng2021learning}
A.~Zeng, X.~Sun, L.~Yang, N.~Zhao, M.~Liu, and Q.~Xu, ``Learning skeletal graph
  neural networks for hard 3d pose estimation,'' in \emph{Proceedings of the
  IEEE/CVF international conference on computer vision}, 2021, pp.
  11\,436--11\,445.

\bibitem{zhang2022uncertainty}
J.~Zhang, Y.~Chen, and Z.~Tu, ``Uncertainty-aware 3d human pose estimation from
  monocular video,'' in \emph{Proceedings of the 30th ACM International
  Conference on Multimedia}, 2022, pp. 5102--5113.

\bibitem{zhou2023dual}
L.~Zhou, Y.~Chen, and J.~Wang, ``Dual-path transformer for 3d human pose
  estimation,'' \emph{IEEE Transactions on Circuits and Systems for Video
  Technology}, 2023.

\bibitem{newell2016stacked}
A.~Newell, K.~Yang, and J.~Deng, ``Stacked hourglass networks for human pose
  estimation,'' in \emph{Computer Vision--ECCV 2016: 14th European Conference,
  Amsterdam, The Netherlands, October 11-14, 2016, Proceedings, Part VIII
  14}.\hskip 1em plus 0.5em minus 0.4em\relax Springer, 2016, pp. 483--499.

\bibitem{sun2019deep}
K.~Sun, B.~Xiao, D.~Liu, and J.~Wang, ``Deep high-resolution representation
  learning for human pose estimation,'' in \emph{Proceedings of the IEEE/CVF
  conference on computer vision and pattern recognition}, 2019, pp. 5693--5703.

\bibitem{kan2023self}
Z.~Kan, S.~Chen, C.~Zhang, Y.~Tang, and Z.~He, ``Self-correctable and adaptable
  inference for generalizable human pose estimation,'' in \emph{Proceedings of
  the IEEE/CVF Conference on Computer Vision and Pattern Recognition}, 2023,
  pp. 5537--5546.

\bibitem{rhodin2018unsupervised}
H.~Rhodin, M.~Salzmann, and P.~Fua, ``Unsupervised geometry-aware
  representation for 3d human pose estimation,'' in \emph{Proceedings of the
  European conference on computer vision (ECCV)}, 2018, pp. 750--767.

\bibitem{li2021hybrik}
J.~Li, C.~Xu, Z.~Chen, S.~Bian, L.~Yang, and C.~Lu, ``Hybrik: A hybrid
  analytical-neural inverse kinematics solution for 3d human pose and shape
  estimation,'' in \emph{Proceedings of the IEEE/CVF conference on computer
  vision and pattern recognition}, 2021, pp. 3383--3393.

\bibitem{burenius20133d}
M.~Burenius, J.~Sullivan, and S.~Carlsson, ``3d pictorial structures for
  multiple view articulated pose estimation,'' in \emph{Proceedings of the IEEE
  conference on computer vision and pattern recognition}, 2013, pp. 3618--3625.

\bibitem{pavlakos2017harvesting}
G.~Pavlakos, X.~Zhou, K.~G. Derpanis, and K.~Daniilidis, ``Harvesting multiple
  views for marker-less 3d human pose annotations,'' in \emph{Proceedings of
  the IEEE conference on computer vision and pattern recognition}, 2017, pp.
  6988--6997.

\bibitem{zhuo2022structural}
Z.~Chen, X.~Zhao, and X.~Wan, ``Structural triangulation: A closed-form
  solution to constrained 3d human pose estimation,'' in \emph{European
  Conference on Computer Vision (ECCV)}, 2022.

\bibitem{WAN2023103830}
X.~Wan, Z.~Chen, and X.~Zhao, ``View consistency aware holistic triangulation
  for 3d human pose estimation,'' \emph{Computer Vision and Image
  Understanding}, vol. 236, p. 103830, 2023.

\bibitem{wandt2019repnet}
B.~Wandt and B.~Rosenhahn, ``Repnet: Weakly supervised training of an
  adversarial reprojection network for 3d human pose estimation,'' in
  \emph{Proceedings of the IEEE/CVF conference on computer vision and pattern
  recognition}, 2019, pp. 7782--7791.

\bibitem{10179252}
M.~T. Hassan and A.~Ben~Hamza, ``Regular splitting graph network for 3d human
  pose estimation,'' \emph{IEEE Transactions on Image Processing}, vol.~32, pp.
  4212--4222, 2023.

\bibitem{9535240}
M.~Li, S.~Chen, Y.~Zhao, Y.~Zhang, Y.~Wang, and Q.~Tian, ``Multiscale
  spatio-temporal graph neural networks for 3d skeleton-based motion
  prediction,'' \emph{IEEE Transactions on Image Processing}, vol.~30, pp.
  7760--7775, 2021.

\bibitem{9009459}
Y.~Cai, L.~Ge, J.~Liu, J.~Cai, T.-J. Cham, J.~Yuan, and N.~M. Thalmann,
  ``Exploiting spatial-temporal relationships for 3d pose estimation via graph
  convolutional networks,'' in \emph{2019 IEEE/CVF International Conference on
  Computer Vision (ICCV)}, 2019, pp. 2272--2281.

\bibitem{vaswani2017attention}
A.~Vaswani, N.~Shazeer, N.~Parmar, J.~Uszkoreit, L.~Jones, A.~N. Gomez,
  {\L}.~Kaiser, and I.~Polosukhin, ``Attention is all you need,''
  \emph{Advances in neural information processing systems}, vol.~30, 2017.

\bibitem{yosinski2014transferable}
J.~Yosinski, J.~Clune, Y.~Bengio, and H.~Lipson, ``How transferable are
  features in deep neural networks?'' \emph{Advances in neural information
  processing systems}, vol.~27, 2014.

\bibitem{yan2022crossloc}
Q.~Yan, J.~Zheng, S.~Reding, S.~Li, and I.~Doytchinov, ``Crossloc: Scalable
  aerial localization assisted by multimodal synthetic data,'' in
  \emph{Proceedings of the IEEE/CVF Conference on Computer Vision and Pattern
  Recognition}, 2022, pp. 17\,358--17\,368.

\bibitem{chang2022maskgit}
H.~Chang, H.~Zhang, L.~Jiang, C.~Liu, and W.~T. Freeman, ``Maskgit: Masked
  generative image transformer,'' in \emph{Proceedings of the IEEE/CVF
  Conference on Computer Vision and Pattern Recognition}, 2022, pp.
  11\,315--11\,325.

\bibitem{shan2022p}
W.~Shan, Z.~Liu, X.~Zhang, S.~Wang, S.~Ma, and W.~Gao, ``P-stmo: Pre-trained
  spatial temporal many-to-one model for 3d human pose estimation,'' in
  \emph{European Conference on Computer Vision}.\hskip 1em plus 0.5em minus
  0.4em\relax Springer, 2022, pp. 461--478.

\bibitem{loop_computing_1999}
C.~Loop and Z.~Zhang, ``Computing rectifying homographies for stereo vision,''
  in \emph{Proceedings. 1999 IEEE Computer Society Conference on Computer
  Vision and Pattern Recognition (Cat. No PR00149)}, vol.~1.\hskip 1em plus
  0.5em minus 0.4em\relax IEEE, 1999, pp. 125--131.

\bibitem{xu_aanet_2020}
H.~Xu and J.~Zhang, ``{AANet}: Adaptive aggregation network for efficient
  stereo matching,'' in \emph{2020 {IEEE}/{CVF} Conference on Computer Vision
  and Pattern Recognition ({CVPR})}.\hskip 1em plus 0.5em minus 0.4em\relax
  {IEEE}, pp. 1956--1965.

\bibitem{xu_attention_2022}
G.~Xu, J.~Cheng, P.~Guo, and X.~Yang, ``Attention concatenation volume for
  accurate and efficient stereo matching,'' in \emph{Proceedings of the
  IEEE/CVF Conference on Computer Vision and Pattern Recognition}, 2022, pp.
  12\,981--12\,990.

\bibitem{kendall_end--end_2017}
A.~Kendall, H.~Martirosyan, S.~Dasgupta, P.~Henry, R.~Kennedy, A.~Bachrach, and
  A.~Bry, ``End-to-end learning of geometry and context for deep stereo
  regression,'' in \emph{Proceedings of the IEEE international conference on
  computer vision}, 2017, pp. 66--75.

\bibitem{sun2018integral}
X.~Sun, B.~Xiao, F.~Wei, S.~Liang, and Y.~Wei, ``Integral human pose
  regression,'' in \emph{Proceedings of the European conference on computer
  vision (ECCV)}, 2018, pp. 529--545.

\bibitem{zheng20213d}
C.~Zheng, S.~Zhu, M.~Mendieta, T.~Yang, C.~Chen, and Z.~Ding, ``3d human pose
  estimation with spatial and temporal transformers,'' in \emph{Proceedings of
  the IEEE/CVF International Conference on Computer Vision}, 2021, pp.
  11\,656--11\,665.

\bibitem{ofli2013berkeley}
F.~Ofli, R.~Chaudhry, G.~Kurillo, R.~Vidal, and R.~Bajcsy, ``Berkeley mhad: A
  comprehensive multimodal human action database,'' in \emph{2013 IEEE workshop
  on applications of computer vision (WACV)}.\hskip 1em plus 0.5em minus
  0.4em\relax IEEE, 2013, pp. 53--60.

\bibitem{ionescu2013human3}
C.~Ionescu, D.~Papava, V.~Olaru, and C.~Sminchisescu, ``Human3. 6m: Large scale
  datasets and predictive methods for 3d human sensing in natural
  environments,'' \emph{IEEE transactions on pattern analysis and machine
  intelligence}, vol.~36, no.~7, pp. 1325--1339, 2013.

\bibitem{makris_robust_2019}
A.~Makris and A.~Argyros, ``Robust 3d human pose estimation guided by filtered
  subsets of body keypoints,'' in \emph{2019 16th International Conference on
  Machine Vision Applications ({MVA})}, pp. 1--6.

\bibitem{ying_rgb-d_2021}
J.~Ying and X.~Zhao, ``Rgb-d fusion for point-cloud-based 3d human pose
  estimation,'' in \emph{2021 {IEEE} International Conference on Image
  Processing ({ICIP})}, pp. 3108--3112, {ISSN}: 2381-8549.

\bibitem{xiao2018simple}
B.~Xiao, H.~Wu, and Y.~Wei, ``Simple baselines for human pose estimation and
  tracking,'' in \emph{Proceedings of the European conference on computer
  vision (ECCV)}, 2018, pp. 466--481.

\bibitem{paszke2017automatic}
A.~Paszke, S.~Gross, S.~Chintala, G.~Chanan, E.~Yang, Z.~DeVito, Z.~Lin,
  A.~Desmaison, L.~Antiga, and A.~Lerer, ``Automatic differentiation in
  pytorch,'' 2017.

\bibitem{he2016deep}
K.~He, X.~Zhang, S.~Ren, and J.~Sun, ``Deep residual learning for image
  recognition,'' in \emph{Proceedings of the IEEE conference on computer vision
  and pattern recognition}, 2016, pp. 770--778.

\bibitem{andriluka14cvpr}
M.~Andriluka, L.~Pishchulin, P.~Gehler, and B.~Schiele, ``2d human pose
  estimation: New benchmark and state of the art analysis,'' in \emph{CVPR},
  June 2014.

\bibitem{moon2018v2v}
G.~Moon, J.~Y. Chang, and K.~M. Lee, ``V2v-posenet: Voxel-to-voxel prediction
  network for accurate 3d hand and human pose estimation from a single depth
  map,'' in \emph{Proceedings of the IEEE conference on computer vision and
  pattern Recognition}, 2018, pp. 5079--5088.

\bibitem{loshchilov2017decoupled}
I.~Loshchilov and F.~Hutter, ``Decoupled weight decay regularization,''
  \emph{arXiv preprint arXiv:1711.05101}, 2017.

\end{thebibliography}
}

\vfill

\end{document}